\documentclass{article}
\usepackage{iclr2027/iclr2027_conference, times}

\usepackage{enumitem}
\usepackage{amsmath,amssymb}
\usepackage{amsthm}

\newtheorem{theorem}{Theorem}

\newtheorem{lemma}{Lemma}

\theoremstyle{definition}
\newtheorem{definition}{Definition}
\newtheorem{assumption}{Assumption}

\theoremstyle{plain}
\usepackage{graphicx}
\usepackage{booktabs}
\usepackage{xcolor}

\usepackage{natbib}
\usepackage{multibib}
\newcites{app}{Appendix References}
\usepackage[colorlinks=true,allcolors=blue]{hyperref}
\graphicspath{{figures/}}

\newcommand{\PKPS}{PKPS\xspace}
\usepackage{xspace}

\title{Benchmark Prediction with Sparse \\ and Unpaired Cached Responses}

\author{Hayden Helm\thanks{Corresponding author: hayden@helivan.io.} \\
Helivan \\
\And
Avijit Ghosh \\
Hugging Face
\And
Ben Johnson \\
Jataware
\And
Carey Priebe \\
Johns Hopkins University
}
\iclrfinalcopy

\begin{document}
\maketitle
\lhead{Preprint.}

\begin{abstract}
Evaluating a generative model on all queries from all available benchmark tasks is prohibitively expensive. 
As such, models are often evaluated on different subsets of tasks and/or different subsets of queries from a given task. The resulting collection of responses and the corresponding (model, task) score matrix are sparse and unpaired. 
In this paper, we introduce the Product Kernel Perspective Space (PKPS), a collection of low-dimensional representations of models based on their embedded responses. 
PKPS generalizes existing methods for representing black-box generative models by comparing responses to similar -- rather than just identical -- queries and is therefore able to utilize sparse and unpaired information. 
Theoretically, we show that benchmark prediction using PKPS is query-efficient relative to methods that only use information from dense and paired portions of the evaluation data. 
Empirically, we show that PKPS-based methods outperform baselines on benchmark prediction across two large benchmark suites.
\end{abstract}

\section{Introduction}
\label{sec:introduction}
Evaluating a generative model is necessary for understanding its capabilities.
As the capabilities of generative models increase \citep{wei2022emergent, bubeck2023sparks},
their applicability expands to more domains---often necessitating the creation of benchmarks for, e.g., mathematics \citep{hendrycks2021math}, law
\citep{guha2023legalbench}, finance \citep{islam2023financebench}, venture capital
\citep{chen2025vcbench}, and agentic tasks \citep{liu2023agentbench}.
The current benchmarking landscape is correspondingly vast: BIG-bench alone contains
more than 200 tasks \citep{srivastava2023beyond}, HELM tracks dozens of scenarios
across multiple releases \citep{helm}, and a recent compilation includes 133 different
benchmarks for frontier models \citep{benchpress}.
As such, a model is rarely evaluated on all benchmarks: of the (model, task) pairs in
the compilation above, only 23\% have a reported score.
Further, benchmarks evolve (e.g., queries are added \citep{chollet2025arcagi2}, refreshed to avoid contamination
\citep{white2024livebench}, or subsampled to reduce cost
\citep{perlitz2024efficient}) and models evaluated on the same benchmark may have responded to different sets of queries.

Benchmark prediction
attempts to alleviate the resource burden of evaluation by using existing information rather than computing scores solely from newly generated
responses.
There are two main benchmark prediction tasks: query-efficient evaluation and
(model, task) score completion.
In query-efficient evaluation, a model responds to a small subset of a task's queries
and its score on the full task is predicted from the resulting responses.
Existing methods for query-efficient evaluation require access to item-level scores \citep{polo2024tinybenchmarks, hofmann2025fluid, li2024active,
bowyer2026efficient} and/or cached embedded responses \citep{dkps_qeff} to queries that all models have responded to and are typically not effective in settings where different models have been evaluated on different sets of queries.
In (model, task) score completion, a model has not responded to any of a task's
queries and its score on the task is predicted from the observed entries of the score
matrix.
Existing methods for score completion use only the aggregate scores and discard the
item-level responses that produced them \citep{benchpress}.

In this paper, we introduce the Product Kernel Perspective Space (\PKPS) to predict
benchmark scores directly from sparse and unpaired collections of responses.
Our contributions can be summarized as follows:
\begin{enumerate}
[wide, labelwidth=1pt, labelindent=12pt]
\setlength{\itemsep}{1pt}\setlength{\parskip}{0pt}
\item \emph{Conceptual:} we unify two prediction
tasks as instances
of a single problem: predicting scores from a sparse, unpaired collection of
responses from previously evaluated models.
\item \emph{Methodological \& theoretical:} we propose the \PKPS, a collection of
low-dimensional representations of models defined on potentially unpaired embedded
responses, and show that benchmark prediction using \PKPS is at least as
query-efficient as methods that only use information from paired responses
(Theorem~\ref{thm:qeff}).
\item \emph{Empirical:} we show
that \PKPS matches the accuracy of the sample score with up to an $8\times$ smaller
query budget and, when ensembled with low-rank matrix completion, reduces score
matrix completion error by up to 28\%. Our empirical evaluation spans two large benchmark suites. 
\end{enumerate}

\begin{table}[t!]
\centering
\caption{MAE of benchmark prediction methods on both suites and tasks (best per column
bolded, with ties both bolded; second best underlined; setup described in
Section~\ref{sec:experiments}):
query-efficiency at budgets $m\in\{1,8\}$ queries per cell at full task coverage, and
completion of missing cells at cohort sizes $n\in\{10,\max\}$ with task coverage
$p_{\mathrm{task}}$ and query depth $p_{\mathrm{query}}$ fixed at $0.5$, except for the
sp.\ columns where $p_{\mathrm{task}}{=}0.2$ and $n{=}10$. ``---'' denotes a setting where a method is not
applicable.}
\label{tab:suite}
\footnotesize
\setlength{\tabcolsep}{3.4pt}\renewcommand{\arraystretch}{0.95}
\begin{tabular}{lccccc ccccc}
\toprule
 & \multicolumn{5}{c}{\textbf{HELM suite} ($18$ tasks, $93$ models)}
 & \multicolumn{5}{c}{\textbf{EEE suite} ($16$ tasks, $45$ models)} \\
\cmidrule(lr){2-6}\cmidrule(lr){7-11}
 & \multicolumn{2}{c}{Query-eff.} & \multicolumn{3}{c}{Completion}
 & \multicolumn{2}{c}{Query-eff.} & \multicolumn{3}{c}{Completion} \\
Method & ($m{=}1$) & ($m{=}8$) & ($n{=}10$) & ($n{=}93$) & (sp.)
       & ($m{=}1$) & ($m{=}8$) & ($n{=}10$) & ($n{=}45$) & (sp.) \\
\midrule
Sample score       & $0.292$ & $\underline{0.092}$ & --- & --- & --- & $0.224$ & $0.081$ & --- & --- & --- \\
IRT                & $0.348$ & $0.343$ & --- & --- & --- & $0.157$ & $0.139$ & --- & --- & --- \\
LRMC               & --- & --- & $0.139$ & $0.105$ & $0.208$ & --- & --- & $0.109$ & $0.096$ & $0.144$ \\
DKPS               & $0.171$ & $0.111$ & --- & --- & --- & $0.116$ & $0.078$ & --- & --- & --- \\
\PKPS              & $\underline{0.137}$ & $0.109$ & $\mathbf{0.120}$ & $\underline{0.104}$ & $\mathbf{0.171}$ & $\underline{0.070}$ & $\underline{0.060}$ & $\mathbf{0.081}$ & $\underline{0.070}$ & $\mathbf{0.108}$ \\
Ensemble           & $\mathbf{0.126}$ & $\mathbf{0.073}$ & $\mathbf{0.120}$ & $\mathbf{0.100}$ & $\underline{0.172}$ & $\mathbf{0.068}$ & $\mathbf{0.050}$ & $\mathbf{0.081}$ & $\mathbf{0.069}$ & $\mathbf{0.108}$ \\
\bottomrule
\end{tabular}
\end{table}

\subsection{Related Work}
\label{sec:related}

\paragraph{Benchmark prediction.}
We decompose benchmark prediction into two lines of prior work: query-efficient
benchmarking and (model, task) score completion.
Methods for query-efficient benchmarking predict a model's score on a full task from
its responses to a small subset of the task's queries---typically by selecting an
informative query subset and extrapolating via item-response theory
\citep{polo2024tinybenchmarks, hofmann2025fluid}, clustering
\citep{vivek2024anchor}, rank-preserving subsampling \citep{perlitz2024efficient},
learned acquisition policies \citep{li2024active}, regression on item-level scores
\citep{bowyer2026efficient}, or regression on model representations
\citep{dkps_qeff}.
All previous methods proposed for query-efficient benchmarking implicitly require
item-level scores or responses on a common set of queries across reference models.

Methods for (model, task) score completion predict entries of the score matrix that
were never evaluated via cross-task predictability \citep{ye2023predictable},
observational scaling laws \citep{ruan2024observational}, or low-rank matrix
completion \citep{benchpress}. All operate exclusively on aggregate scores and
discard information in the item-level scores or responses.
Our proposed method uses item-level information from sparse and unpaired cross-task evaluation
data to improve on methods from both literatures.

\paragraph{Low-dimensional representations of models.}
Our approach to sparse and unpaired benchmark prediction first constructs low-dimensional representations of the models before learning a simple regressor in the representation space.
Existing approaches to represent models vary by access regime.
For example, representations can be induced by comparing internal activations
\citep{duderstadt2023comparing, huh2024platonic} and/or weights
\citep{chen2025weights} in the white-box setting.
Evaluators do not necessarily have white-box access and instead rely on black-box methods such as
the Data Kernel Perspective Space (DKPS) \citep{helm2024tracking,helm2025inference}, which constructs low-dimensional representations based on distances between matrices of embedded responses on a paired set of queries.
Most relevant to our setting, \citet{dkps_qeff} regress benchmark scores of
previously evaluated models on DKPS representations and prove that the predictions are query-efficient relative to the sample score.
The proposed Product Kernel Perspective Space generalizes this construction with a kernel on the query space, enabling comparisons in the sparse and unpaired regime.

\section{Benchmark Prediction}
\label{sec:setting}
For our purposes, a model is a random mapping
$f:\mathcal{Q}\to\mathcal{X}$ from a query space to a response space.
A task is a pair $(Q^{(t)}, s^{(t)})$, where $Q^{(t)}\subseteq\mathcal{Q}$ is a
collection of queries with $M^{(t)}:=|Q^{(t)}|$ and $s^{(t)}$ is a scoring function
mapping a model and a query collection to a nonnegative real number. 
Model $i$'s true score on task
$t$ is $y_i^{(t)}:=s^{(t)}(f_{i},Q^{(t)}) \in \mathbb{R}_{\ge 0}$.

Evaluation data is indexed by (model, task, query) triples: for $n$ models
$f_1,\dots,f_n$ and $T$ tasks $Q^{(1)},\dots,Q^{(T)}$, the response of model $i$ to
query $q\in Q^{(t)}$ may or may not have been generated and scored.
Let $Q_{i}^{(t)}\subseteq Q^{(t)}$ be the queries for which model $i$'s response (and score) is
observed, with $m_{i}^{(t)}:=|Q^{(t)}_{i}|$ and $Q_i:=\cup_{t}Q_i^{(t)}$.
The corresponding sample score
$\tilde{y}_{i}^{(t)}:=s^{(t)}(f_{i},Q_{i}^{(t)})$ is an estimate of
$y_{i}^{(t)}$.

Evaluation data is \emph{sparse} when few models have
$m_{i}^{(t)}>0$ for a given task. It is \emph{unpaired} when the sizes of cross-model query set intersections 
$m_{ii'}^{(t)}:=|Q_{i}^{(t)}\cap Q_{i'}^{(t)}|$ are small. We let
$\rho_{ii'}^{(t)}:=m_{ii'}^{(t)}/m_{i}^{(t)}$ be the ratio of shared queries for a
given model pair and note that $\rho_{ii'}^{(t)}\approx1$ for all pairs in highly curated
leaderboards and $\rho_{ii'}^{(t)}\approx0$ in deployed evaluation settings.
The goal of benchmark prediction is to estimate a model's score from the available
evaluation data.

The two tasks introduced in \S\ref{sec:introduction} are the two regimes of the
observation count $m_i^{(t)}$:
\begin{enumerate}[wide, labelwidth=1pt, labelindent=12pt]
\setlength{\itemsep}{1pt}\setlength{\parskip}{0pt}
\item \emph{Query-efficient evaluation} is the case
with $m_{i}^{(t)}>0$. Responses or scores from other models on task $t$ are
typically available.
\item \emph{(Model, task) score matrix completion} is the case with $m_{i}^{(t)}=0$.
No responses from the target (model, task) slice exist and prediction
must be based on model $i$'s responses to other tasks and on responses from
other models evaluated on task $t$.
\end{enumerate}

\section{The Product Kernel Perspective Space}
\label{sec:method}

We estimate $y_i^{(t)}$ from sparse and unpaired
evaluation data by constructing low-dimensional
representations of the models based on an appropriately defined pairwise distance
matrix.
We assume access to the responses for each observed (model, task, query) triple and
to two embedding functions $g_{Q}$ and $g_{R}$ that map
queries and responses to $\mathbb{R}^{p_{Q}}$ and $\mathbb{R}^{p_{R}}$, respectively.
Let $q_{1},q_{2},\hdots, q_{M} $ enumerate the observed queries and let
$x_{ij}$ denote the embedded response $f_{i}(q_{j})$ and $u_{j}$ denote the embedded
query $q_{j}$.

Given a query kernel
$k_Q:\mathbb{R}^{p_{Q}}\times\mathbb{R}^{p_{Q}}\to\mathbb{R}_{\ge0}$ and a response
kernel $k_R:\mathbb{R}^{p_{R}}\times\mathbb{R}^{p_{R}}\to\mathbb{R}_{\ge0}$, we
compare models $i$ and $i'$ by computing the pairwise affinity matrix $ A $:
\begin{equation}
A_{ii'} = \frac{\sum_{j\in Q_i}\sum_{l\in Q_{i'}} k_Q(u_j,u_l)\, k_R(x_{ij},x_{i'l})}
              {\sum_{j\in Q_i}\sum_{l\in Q_{i'}} k_Q(u_j,u_l)}.
\label{eq:A}
\end{equation}
Each entry is normalized by its own query-kernel mass, so models that answered
different numbers of queries, or entirely different queries, remain comparable.
We use the induced squared dissimilarity $D^2_{ii'}=A_{ii}+A_{i'i'}-2A_{ii'}$ to
capture model (dis)similarity.
Finally, the Product Kernel Perspective Space representations of the
models---$\widehat{\psi}_{1},\hdots,\widehat{\psi}_{n}$---are obtained by classical
multidimensional scaling (CMDS) of $D^2$ \citep{torgerson1952mds, borg2005modern}.
That is, let $B=-\tfrac12 J D^2 J$ with $J=I-\tfrac1n\mathbf{1}\mathbf{1}^\top$, and let
$\lambda_1\ge\dots\ge\lambda_n$ and $v_1,\dots,v_n$ denote its eigenvalues and
eigenvectors. Then
\begin{equation}
\widehat{\Psi}=\big[\hat\psi_1,\dots,\hat\psi_n\big]^\top
=\big[v_1,\dots,v_d\big]\,\operatorname{diag}\!\big((\lambda_1)_+^{1/2},\dots,(\lambda_d)_+^{1/2}\big),
\label{eq:mds}
\end{equation}
so that $\widehat{\Psi}\widehat{\Psi}^\top$ is the best rank-$d$ positive semidefinite
approximation of $B$.
When $D^2$ is a matrix of squared Euclidean distances in dimension at most $d$, CMDS
recovers the corresponding configuration up to a rigid motion \citep{Dokmanic_2015}.

\begin{figure}[t]
\centering
\includegraphics[width=\textwidth]{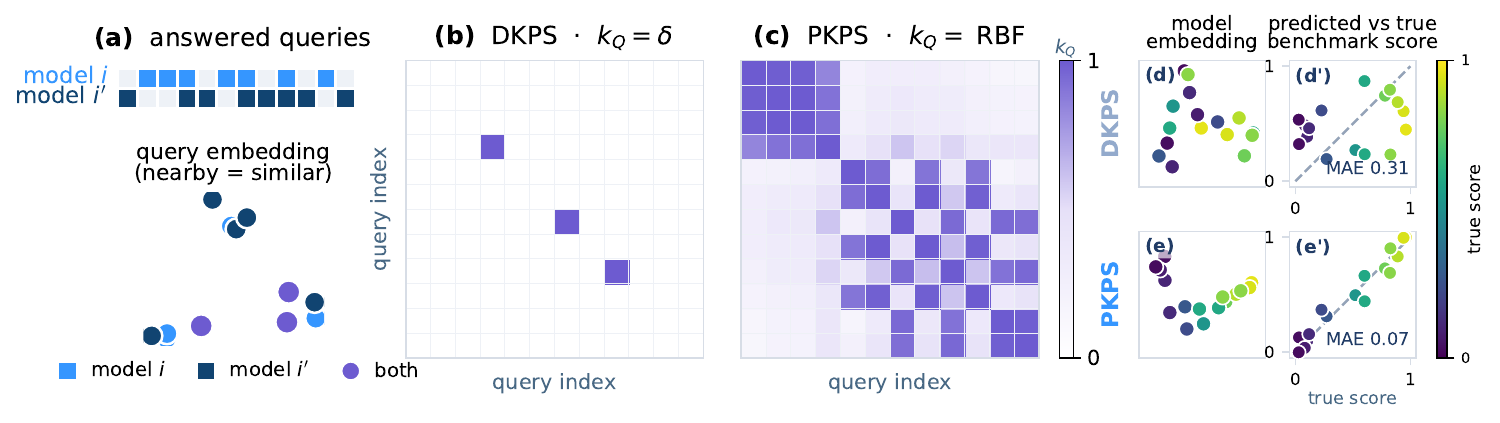}
\caption{An illustration of the difference between the Product Kernel Perspective
Space (\PKPS) and the Data Kernel Perspective Space (DKPS).
\textbf{(a)} Top: two models respond to different sets of queries. Bottom: queries
represented as vectors, colored by which model has responded to them. \textbf{(b)} The query kernel used by DKPS. 
Only
information from queries both models have responded to is included. \textbf{(c)} The query kernel used by \PKPS weighs query similarity in the embedding space. \textbf{(d--e$'$)} A benchmark prediction
example. Left: each dot is a model colored by its benchmark score. Right: true benchmark score versus predicted benchmark score. Restricted to the rare shared queries, DKPS (top) is noisy and correlates poorly
with benchmark score (MAE $0.31$). \PKPS (bottom) leverages unpaired queries and is highly
correlated with true score (MAE $0.07$).}
\label{fig:concept}
\end{figure}

We note that the definition of the \PKPS strictly generalizes previous work by
replacing the identity query kernel ($k_{Q}(u,u')=\mathbf{1}\{u=u'\}$) with a generic
query kernel.
Importantly, for any strictly positive $k_Q$, such as RBF, $D^2_{ii'}$ includes information from unpaired responses and Eq.~\eqref{eq:A}
is therefore always defined when $ |Q_{i}|, |Q_{i'}| \ge 1 $.
We highlight the difference between the previous approaches and the \PKPS in Figure~\ref{fig:concept}.

\subsection{Concentration of \PKPS representations}
\label{sec:theory}

We next establish theoretical properties of the \PKPS necessary for demonstrating
its efficiency in sparse and unpaired benchmark prediction.
The main result of this section establishes that the representations
concentrate around their population counterparts at a rate governed by the overlap of
the query distributions.
We model the evaluation data of Section~\ref{sec:setting} as
sampled (task, query, response) triples: conditional on model $i$, each triple
independently draws a task, then a query from the task, and finally the model's
response (formalized in Definition~\ref{def:pop} of Appendix~\ref{app:proof}).
We formally define the population \PKPS in Definition~\ref{def:pop} of
Appendix~\ref{app:proof}; broadly, it is defined by replacing the sums of
Eq.~\eqref{eq:A} with expectations.

Our concentration result requires bounded query and response kernels, bounded
embedded responses, a positive cross-model query kernel mass, and a gap after the
$d$-th eigenvalue of the population double-centered matrix (see
Assumptions~\ref{ass:bound}--\ref{ass:spec}, Appendix~\ref{app:proof}).
The required boundedness and the spectral gap are standard in spectral-embedding
analyses \citep{yu2015useful, acharyya2024consistent, acharyya2025concentration}.
The atypical requirement in our setting is the positive cross-model query kernel
mass, though this holds trivially under a strictly positive $k_Q$ and does not assume that any individual query is shared.

Let $(\psi_1,\dots,\psi_n)$ denote the population \PKPS, $\kappa=\inf_{i,i'}\mathbb{E}[k_Q(u,u')]$ the cross-model
query kernel mass, $\lambda_d$ the $d$-th largest eigenvalue of the population
double-centered matrix $B^*$, $\Delta=\lambda_d-\lambda_{d+1}$ its eigengap, and
$\beta=\|B^*\|_{\mathrm{op}}/\Delta$.
\begin{theorem}[Concentration of \PKPS representations]
\label{prop:conc}
Suppose Assumptions~\ref{ass:bound}--\ref{ass:spec} hold and that every model has
responded to at least $m$ queries. Then for any
$\delta\in(0,1)$, with probability at least $1-\delta$ there is an orthogonal $W$ such
that
$\max_i\|\hat\psi_i-W\psi_i\|\le
C\kappa^{-2}\lambda_d^{-1/2}(1+\beta)\,n\sqrt{\log(4n^2/\delta)/(2m)}$,
where $C$ depends only on the kernel bounds.
\end{theorem}

Theorem~\ref{prop:conc} states that the \PKPS representations concentrate at the rate in $ m $ enjoyed by fully paired DKPS \citep{acharyya2025concentration} without requiring any shared queries.
The price of unpairedness is a factor of $\kappa^{-2}$, governed by
the overlap of the query distributions.
The full proof is in Appendix~\ref{app:proof}.

\subsection{Inference in the \PKPS}
\label{sec:qeff}
We next describe benchmark prediction in the \PKPS and establish its query
efficiency.
Inference using the \PKPS is a two-step statistical problem: first embed the
models (Eq.~\eqref{eq:mds}), then train a decision function that uses the
representations as proxies for the models.

Fix a task $t$ and let $f_{0},f_{1},\hdots,f_{n}$ be i.i.d.\ draws from a model
distribution.
Our goal is to predict the true score of $f_{0}$ on task $t$,
$y_{0}:=s^{(t)}(f_{0},Q^{(t)})$.
The (model, task) pairs with $m_{i}^{(t)}>0$ and model $ f_{0}$'s responses from tasks $ t' \neq t $ provide information potentially
relevant to predicting $y_{0}$.

We embed all $n{+}1$ models jointly via Eq.~\eqref{eq:mds}, fit a regressor $\hat h^{(t)}:\mathbb{R}^{d}\to\mathcal{Y}$ on
the observed pairs $\{(\hat\psi_{i},\tilde{y}_{i}^{(t)})\}_{i=1}^{n}$, and
predict $\hat y^{(t)}_{0}=\hat h^{(t)}(\hat\psi_{0})$.
We note that Eq.~\eqref{eq:A} uses no
score information, so the target model is embedded alongside the reference models.
Out-of-sample extensions exist (see, e.g., \citealp{trosset2008out, 10.3389/fnhum.2022.930291}).

\citet{dkps_qeff} established that nearest neighbor regression using representations
from paired DKPS is query efficient relative to the sample score.
The main result of this section shows that benchmark prediction with \PKPS is
query-efficient relative to paired DKPS.
The first part of our result requires only bounded scores
(Assumption~\ref{ass:noise}). The second part additionally requires that no two
query designs share an atom, so that independently sampled query sets are disjoint
almost surely (Assumption~\ref{ass:gap}), that models and their caches are drawn
independently (Assumption~\ref{ass:iid}), and a spectral gap for the fixed-budget
analogue of the population double-centered matrix (Assumption~\ref{ass:specm}).
All assumptions are stated in Appendix~\ref{app:proof}.
In the following, MSE denotes mean squared error and
$m$ denotes a per-model query budget with $\min_{i}|Q_{i}|\ge m$.
We assume the query kernel is from a family
of bounded kernels indexed by a bandwidth $\sigma$ whose
$\sigma\to0$ limit is such that $ k_{0}(u, u') > 0 $ if $ u = u' $ and $ 0 $
otherwise (e.g., $k_{\sigma}(u,u')=\exp(-\|u-u'\|^{2}/2\sigma^{2})$).

\begin{definition}[Query efficiency; following \citealp{dkps_qeff}]
\label{def:qeff}
Fix a per-model query budget $m$ and let $h_n$ denote a score estimator using $n$
reference models. The sequence $(h_{1}, h_{2}, \hdots)$ is \emph{query-efficient
relative to} $(h'_{1}, h'_{2}, \hdots)$ if there exists $N$ such that
$\mathrm{MSE}(h_n)\le\mathrm{MSE}(h'_n)$ with high probability for all $n\ge N$,
with the inequality strict for at least one $n^{*}\ge N$.
\end{definition}
We write $h_{n,\sigma}^{(K)}$ for an estimator that uses a $K$-nearest-neighbor
regressor on the \PKPS constructed with a query kernel with bandwidth $\sigma$.
Since $k_{0}$ only puts positive mass on identical queries, $h_{n,0}^{(K)}$ is
$K$-nearest-neighbor regression on the paired DKPS.
Let $\hat h_{n}^{(K)}$ denote the cross-validated \PKPS estimator (specified in
Appendix~\ref{app:proof}), where the
bandwidth is selected on held-out reference models over a finite grid containing
$\sigma=0$.
A nonzero bandwidth is chosen only when the estimated improvement
exceeds the validation confidence radius.
Finally, let $I_{m}(\sigma)$ denote the variance of the target score explained by the
population representation of the target model computed from its $m$ cached responses
(Definition~\ref{def:info}).
\begin{theorem}[\PKPS is query efficient relative to paired DKPS]
\label{thm:qeff}
Fix any per-model query budget $m\ge1$ and suppose the reference scores are observed
exactly.
(i) Under Assumption~\ref{ass:noise}, for every pairing fraction $\rho\in[0,1]$ and
every $K$, with high probability for all $n$ sufficiently large,
$\mathrm{MSE}\big(\hat h_{n}^{(K)}\big)\le\mathrm{MSE}\big(h_{n,0}^{(K)}\big)$.
(ii) Suppose Assumptions~\ref{ass:noise}, \ref{ass:bound}, \ref{ass:gap},
\ref{ass:iid}, and~\ref{ass:specm} hold and the grid contains a bandwidth
$\sigma^{\star}>0$ with $I_{m}(\sigma^{\star})>0$. Then for every
$0<c<I_{m}(\sigma^{\star})$ and every $K_{n}\to\infty$ with $K_{n}/n\to0$, with high
probability for all $n$ sufficiently large,
$\mathrm{MSE}\big(\hat h_{n}^{(K_{n})}\big)\le\mathrm{MSE}\big(h_{n,0}^{(K_{n})}\big)-c$.
In particular, $(\hat h_{n}^{(K_{n})})$ is query-efficient relative to
$(h_{n,0}^{(K_{n})})$ at every budget $m$ with $I_{m}(\sigma^{\star})>0$.
\end{theorem}
Theorem~\ref{thm:qeff} states that, at any fixed budget, \PKPS is never worse than
paired DKPS at any pairing level. When query sets are disjoint
(Assumption~\ref{ass:gap}), paired DKPS can do no better than predicting the mean
score, while \PKPS improves on it by the score variance $I_{m}(\sigma^{\star})$ that
its budget-$m$ representation explains. Some condition of this kind is needed: if a
model's cached responses carry no information about its score, no estimator built
from them improves on the mean.
We note that query-efficiency is transitive and that Theorem~\ref{thm:qeff}(i),
together with Theorem 2 of \citet{dkps_qeff}, implies that $\hat h_{n}^{(K)}$ is
query-efficient relative to the sample score.

The same machinery covers (model, task) score completion ($m_i^{(t)}=0$): since
Eq.~\eqref{eq:A} uses only responses, a model that has answered none of task $t$'s
queries is embedded from its responses to other tasks, and the regression transfers
the scores of reference models evaluated on $t$.
We evaluate both query-efficiency and completion empirically in Section~\ref{sec:experiments}.

\section{Experiments}
\label{sec:experiments}
We evaluate \PKPS on synthetic data and on two large benchmark suites.

\paragraph{Synthetic data.} We generate $n$ models with latent ability vectors
$v_i \in \mathbb{R}^{5}$ and $T$ tasks with latent vectors $t_k \in \mathbb{R}^{5}$,
scores $v_i\cdot t_k+\epsilon$ with additive noise $ \epsilon $, and queries drawn near each task center ($u_j = t_k +
\eta_j$ with isotropic noise $\eta_j$).
Each model answers $m$ queries per task. 
We control unpairedness via
$\rho=m_{ii'}/m$ and sparsity via a probability of observing a task
$p_{\mathrm{task}}$ and, conditioned on observing a task, a probability of answering
a query $p_{\mathrm{query}}$.

\paragraph{Benchmark suites.}
We consider two large benchmark suites.
The first consists of $18$ tasks from HELM-Lite \citep{helm} 
-- $7$ math subjects (MATH, \citealp{hendrycks2021math}), $5$ translation pairs
(WMT-14, \citealp{bojar2014wmt}), medical multiple choice
(MedQA, \citealp{jin2021medqa}), and $5$ legal subjects
(LegalBench, \citealp{guha2023legalbench}).
We consider $93$ models that have been evaluated on all $ 18 $ tasks and induce sparsity and unpairedness by subsampling each model's observed queries.

The second suite is built from the Every Eval Ever datastore \citep{eee2026}, a
community-maintained collection of item-level evaluation results covering $36$ benchmarks and $165$ models (and growing).
The store exhibits the sparsity and unpairedness that motivate this paper:
only $15\%$ of (model, benchmark) pairs have been evaluated and $59\%$ of model
pairs share no benchmark at all.
Within a benchmark, pairedness is bimodal: on $26$ of $33$ tasks,
models answered essentially identical item sets (mean $\rho_{ii'}>0.99$), while the
remaining seven -- all arena-style \citep{chiang2024chatbot} -- are severely
unpaired (mean $\rho_{ii'}$ between $0.27$ and $0.66$).
We study a dense 45 model and 16 task submatrix so that true scores are available and
sparsity and pairedness can be controlled.
The tasks include MATH-MC
levels $1$--$5$ \citep{hendrycks2021math}, GSM-MC \citep{cobbe2021gsm8k},
GPQA-Diamond \citep{rein2023gpqa}, nine categories from JudgeBench
\citep{tan2024judgebench} and RewardBench~2 \citep{malik2025rewardbench2}. 

Both suites were chosen because they include item-level responses -- most large
public benchmark leaderboards only include scores.
Details describing the HELM and EEE data such as the included models, included
tasks, the number of queries per task, etc.\ are provided in
Appendices~\ref{app:details} and~\ref{app:eee}, respectively.

\paragraph{Methods and baselines.} We evaluate \PKPS on both query-efficiency and
(model, task) score completion.
Unless otherwise stated we embed all responses and queries into a $3072$-dimensional
space via \texttt{gemini-embedding-001} \citep{lee2025gemini}, apply PCA separately to the collection of embedded queries and embedded responses, and use an RBF kernel on the
reduced query space and a linear kernel on the reduced response space.
Given the \PKPS pairwise dissimilarity matrix, we embed the models in a $d = 8$ space
via Eq.~\eqref{eq:mds} and use $K$-nearest neighbor regression with $K = 5$.

Depending on the task we include different baselines. Each produces an estimate of
$y_{i}^{(t)}$:
\begin{enumerate}[wide, labelwidth=1pt, labelindent=12pt]
    \item \emph{Query-efficiency}. We compare \PKPS to the sample score (i.e.,
    $\tilde{y}_{i}^{(t)}$), paired DKPS \citep{dkps_qeff}, Item Response Theory
    (``IRT'', \citealp{polo2024tinybenchmarks}) and a convex combination
    of \PKPS and sample score (i.e.,
    $\alpha\,\tilde{y}_{i}^{(t)}+(1-\alpha)\,\hat{y}^{(t)}_{i,\mathrm{PKPS}}$).
    \item \emph{Score completion.} We compare \PKPS to Low-Rank Matrix Completion
    (``LRMC'', \citealp{benchpress}) and a convex combination of LRMC and \PKPS
    (i.e., $\alpha\,\hat{y}^{(t)}_{i,\mathrm{LRMC}}+(1-\alpha)\,\hat{y}^{(t)}_{i,\mathrm{PKPS}}$).
\end{enumerate}
Both the query kernel RBF bandwidth and the convex coefficient for the ensemble are
selected via cross-validation on the observed data.
Following \citet{benchpress}, the completion methods operate on a pre-processed score matrix:
observed scores are logit-transformed, standardized per task, and centered by a
two-way (model, task) additive bias.
Following \citet{dkps_qeff}, we use the leave-one-family-out protocol where all models from the
target model's developer are excluded. 
As such, the reported errors are conservative.
We report average mean absolute error (MAE) of the predicted score versus the true
score for every setting over $16$ seeds ($50$ on synthetic data).
Additional experimental details are provided in Appendix~\ref{app:details}.

\subsection{Results}
\label{sec:results}

\begin{figure}
    \centering
    \includegraphics[width=0.9\linewidth]{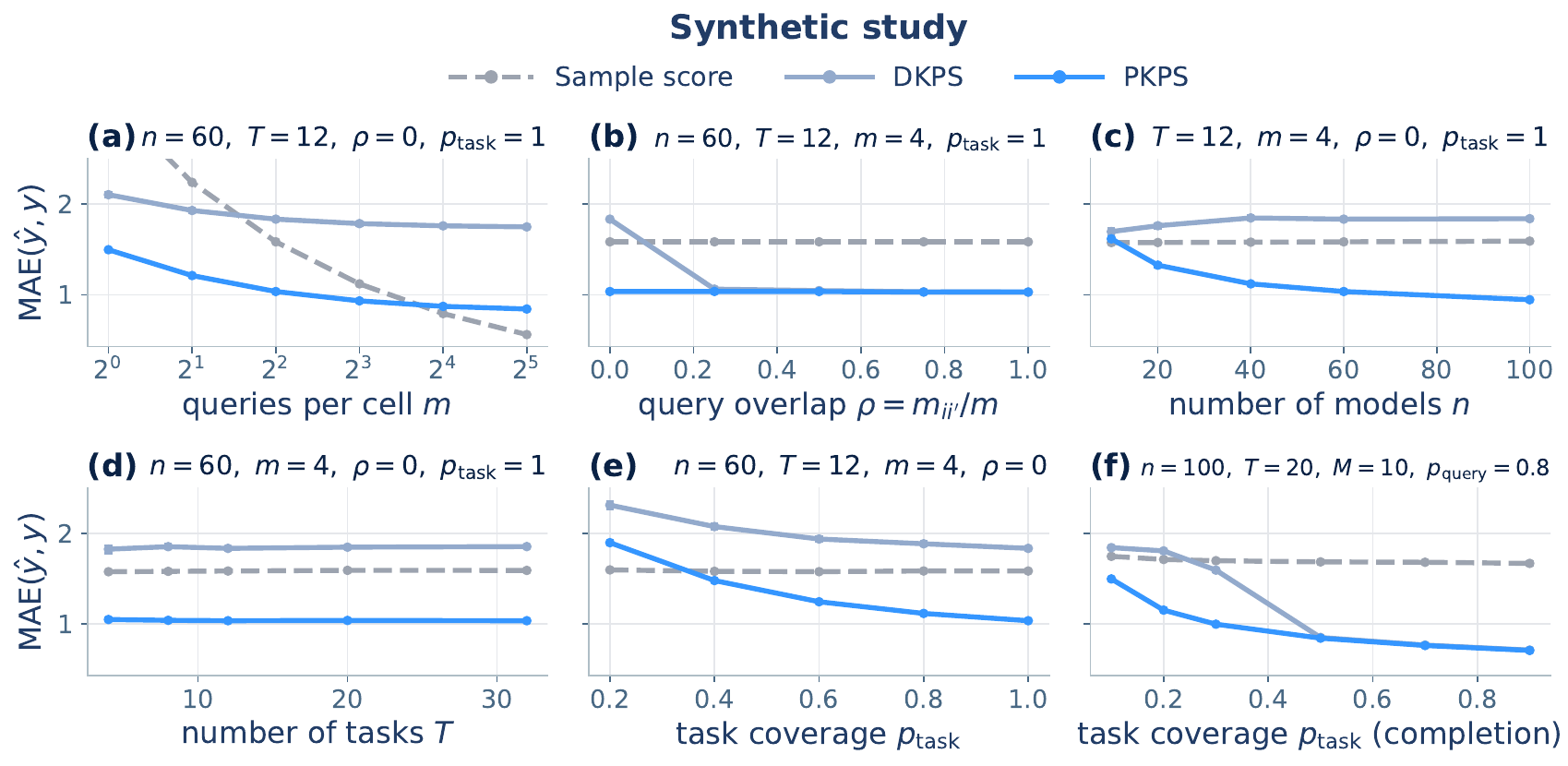}
    \caption{MAE on the synthetic benchmark ($\pm1$ SEM).
\textbf{(Panels a--e)} Query-efficient evaluation where every observed cell is measured with
$m$ queries. We vary the budget
$m$ (panel a), the cross-model overlap $\rho$ (panel b), the cohort size $n$ (panel c), the
number of tasks $T$ (panel d), and the task coverage $p_{\mathrm{task}}$ (panel e), at unpaired
designs ($\rho{=}0$) except where $\rho$ is swept.
\textbf{(panel f)} Score completion of held-out cells as a function of task coverage.}
    \label{fig:synthetic}
\end{figure}

\paragraph{\PKPS is query-efficient on the synthetic data.}
Figure~\ref{fig:synthetic} shows the MAE for \PKPS, DKPS, and the sample score for estimating a model's true score under
the query-efficient evaluation protocol.
Every observed cell of the score matrix is calculated with a budget of
$m$ queries.
We vary the query budget  (panel a), the pairwise query overlap (panel b), the number
of models (panel c), the number of tasks (panel d), and the task coverage (panel e).
Panel (f) studies score completion, where the targets are held-out cells and the
score-only baseline is the task mean.

\PKPS outperforms DKPS everywhere and the sample score at small query budgets (panel a).
Panel (b) is the empirical counterpart of Theorem~\ref{thm:qeff}.
At $\rho=0$, DKPS is worse than the sample score, while a
single shared query ($\rho=0.25$ at $m=4$) restores it to \PKPS's level.
\PKPS is flat in $\rho$ as it only requires distributional overlap.
As $n$ grows for a fixed sparsity and pairedness, \PKPS can take advantage of
additional reference models while DKPS cannot (panel c).
Task coverage behaves similarly (panel e) and the number of tasks matters little for
either method (panel d) --- though this observation likely hinges on the ratio between task-level noise and query-level noise.
In the completion panel (panel f), \PKPS outperforms DKPS when observed cells are scarce
and they are matched for moderate to large coverage.

\begin{figure}[t]
\centering
\includegraphics[width=\textwidth]{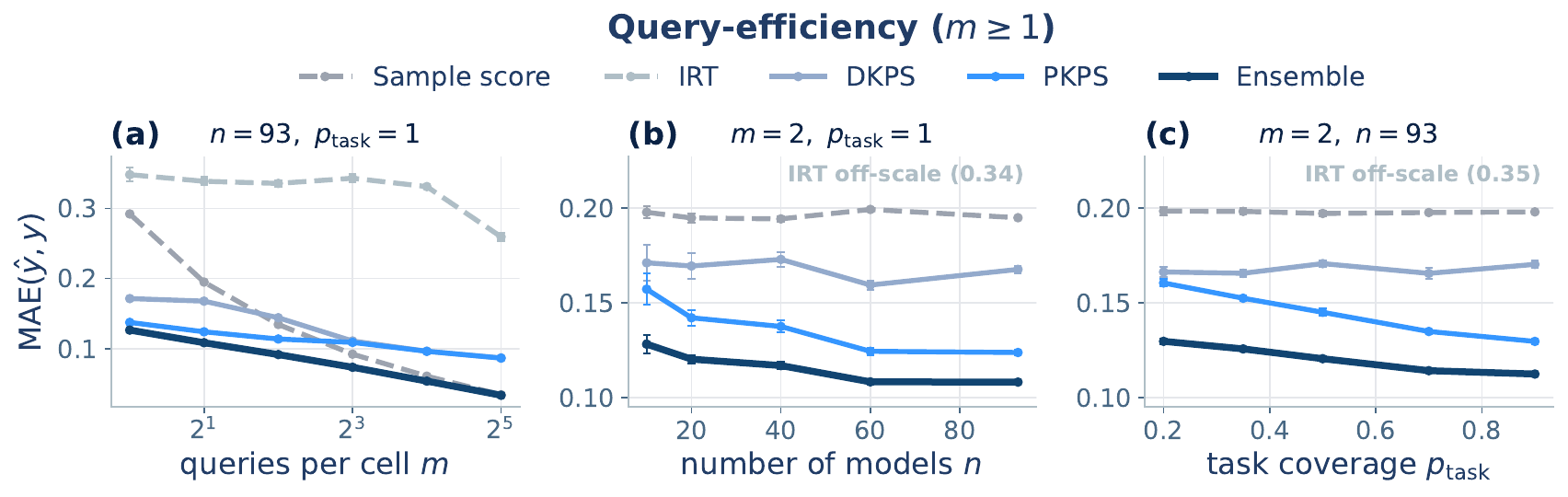}\\[2pt]
\includegraphics[width=\textwidth]{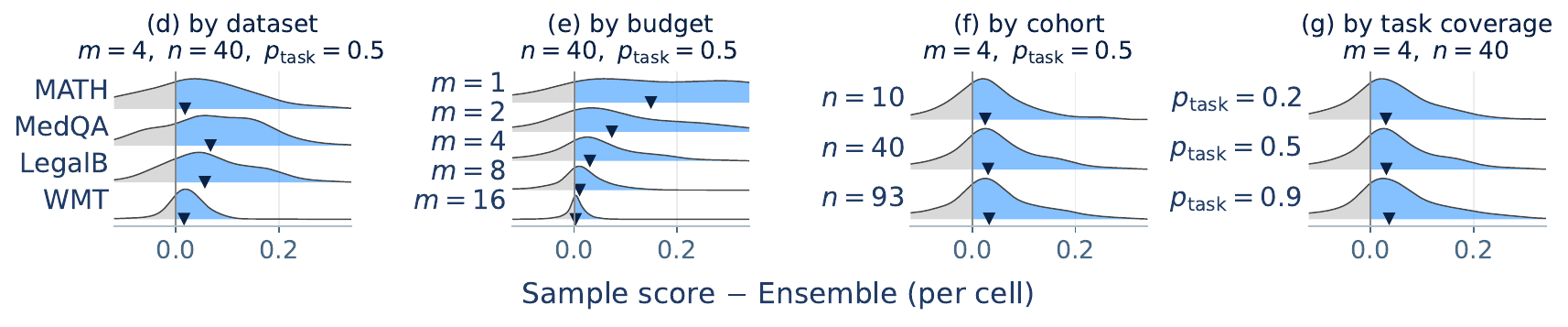}
\caption{Query-efficiency on the HELM suite.
\textbf{(a--c)} MAE against the true full-evaluation score when each model answers $m$ unpaired queries per observed cell
($\pm1$ SEM), as a function of budget $m$, cohort size $n$, and task coverage
$p_{\mathrm{task}}$.
\textbf{(d--g)} Per-cell difference in MAE between
Sample score and Ensemble (positive difference favors Ensemble) by dataset, budget, cohort, and
coverage. $\blacktriangledown$ is the mean for a given setting.}
\label{fig:qe}
\end{figure}

\paragraph{\PKPS-based methods dominate query-efficient evaluation.}
Figure~\ref{fig:qe}a--c shows the MAE for \PKPS, DKPS, IRT, the sample score, and
the ensemble on the HELM suite.
Each model answers its own independently sampled queries.
At $m{=}1$, \PKPS attains $0.137$ and the ensemble $0.126$ versus $0.292$ for the sample score (Wilcoxon $p{=}8\times10^{-6}$ on average task error differences, better on all $18$ tasks), so \PKPS with one query matches the sample score at $ m = 4 $ (panel a).
As the budget grows, the sample score's noise falls below the representations'
residual bias and it eventually overtakes them.
The ensemble follows the better of its two components
throughout (panel a).
\PKPS improves with the number of reference models
(panel b; $0.157\!\to\!0.124$) and with the task coverage (panel c; $0.161\!\to\!0.130$),
while DKPS is flat.
IRT is poorly identified without a shared item block and is worse than the sample
score throughout.

\paragraph{The gains hold across settings.}
Figure~\ref{fig:qe}d--g shows the distribution of the per-cell difference between
the sample score's and the ensemble's absolute errors, by dataset (panel d), budget (panel e),
cohort (panel f), and task coverage (panel g).
The ensemble outperforms the sample score on $69\%$ of cells ($78\%$ at $m{=}1$)
and in the mean of every row.
The gains are largest on the rich-response datasets (panel d) and at the smallest budgets
(panel e), where a single response carries the most information, and they shrink toward
zero as the budget grows.
The mean gain is positive at every cohort size (panel f) and every coverage (panel g).

\begin{figure}[t]
\centering
\includegraphics[width=\textwidth]{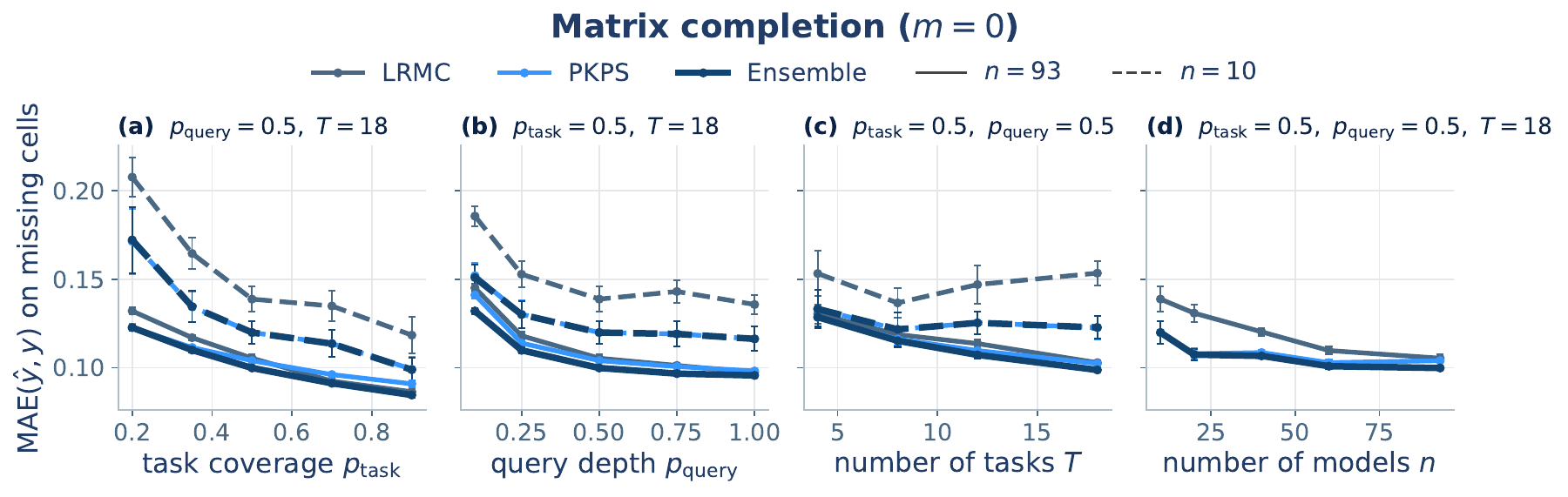}\\[2pt]
\includegraphics[width=\textwidth]{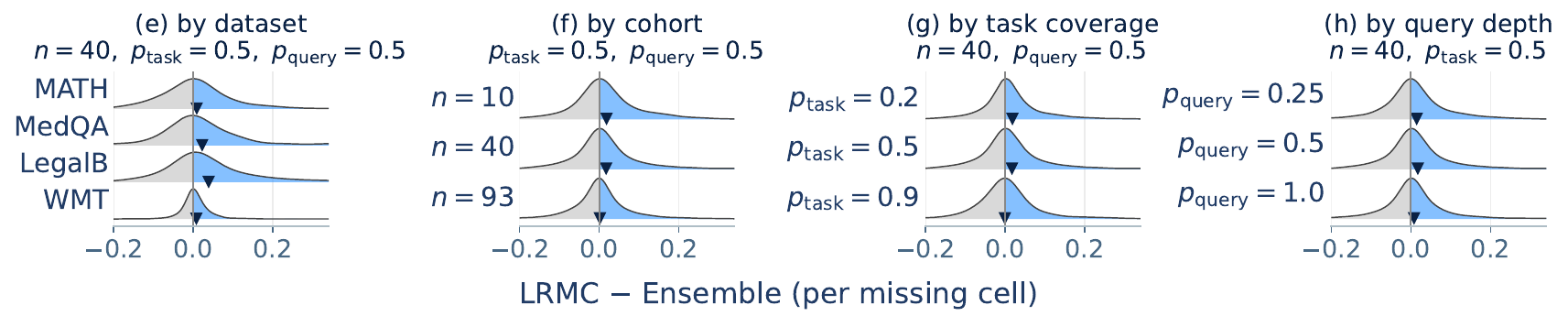}
\caption{Score completion on the HELM suite. \textbf{(a--d)} MAE on missing cells
($\pm1$ SEM) when a fraction $p_{\mathrm{task}}$ of (model, task) cells is observed at
query depth $p_{\mathrm{query}}$, as a function of task coverage, query depth, number of
tasks, and cohort size. \textbf{(e--h)} Per-cell difference in MAE between LRMC and Ensemble (positive difference favors Ensemble) by dataset, cohort, coverage, and query depth. $\blacktriangledown$ is the mean for a given setting.}
\label{fig:completion}
\end{figure}

\paragraph{The same representations complete missing cells in the score matrix.}
Figure~\ref{fig:completion}a--d shows the MAE for LRMC, \PKPS, and the ensemble on
missing cells when a fraction $p_{\mathrm{task}}$ of (model, task) cells is observed
at query depth $p_{\mathrm{query}}$, at cohort sizes $n{=}93$ (solid) and $n{=}10$
(dashed).
No responses from a missing cell exist, so the sample score is unavailable.
The ensemble matches or outperforms LRMC at every operating point, and the margin is
largest when observations are scarce: at the small cohort it is more accurate at
every task coverage (panel a), query depth (panel b), and number of tasks (panel c)
($0.120$ versus $0.139$; Wilcoxon
$p{=}1\times10^{-3}$, better on $15$ of $18$ tasks).
LRMC requires many models to estimate its factors: with $n{=}10$, additional tasks
do not close the gap (panel c), and sweeping the number of reference models shows the ensemble is accurate
from $n{=}20$ while LRMC approaches it only near the full cohort (panel d).
At the full cohort the methods converge and the ensemble remains best on average
($0.100$ versus $0.105$), although the per-task difference is not significant
(better on $9$ of $18$ tasks).

\paragraph{The completion gains are modest but consistent.}
Figure~\ref{fig:completion}e--h shows the distribution of the per-missing-cell
difference between LRMC's and the ensemble's absolute errors, by dataset (panel e),
cohort (panel f), coverage (panel g), and query depth (panel h).
The win is smaller than in the query-efficiency task but the mean gain is positive in every row.

\paragraph{The results are qualitatively similar on the Every Eval Ever suite.}
The second suite shows the same behavior with larger margins (lever sweeps in
Appendix~\ref{app:eee}).
At $m{=}1$, \PKPS attains less than a third of the sample score's error ($0.070$ versus
$0.224$), below the accuracy the sample score reaches at $m{=}8$ ($0.081$).
On missing cells, the ensemble outperforms LRMC everywhere, including
the full cohort (Wilcoxon $p{=}3\times10^{-5}$, better on every task), which is not significant on HELM.
Table~\ref{tab:suite} shows the performance of all methods on both tasks and both suites.
A \PKPS-based method is the best in every column.

\paragraph{The results are insensitive to representation and pipeline choices.} Varying
the MDS dimension over $4$--$24$ moves \PKPS's MAE by less than $0.02$, and the
cross-validated bandwidth matches the best fixed choice (Appendix~\ref{app:sens}).
Ablations of the regression method, the response kernel, the ensemble weighting rule, and
the incidental query overlap are also in Appendix~\ref{app:sens}.
Our observations are generally robust across these choices.

\section{Discussion}
\label{sec:discussion}
In this paper we introduced the \PKPS, a collection of low-dimensional
representations of models defined on sparse and unpaired collections of embedded
responses --- a realistic setting outside of highly curated leaderboards and evaluation collections.
We prove that the \PKPS is query-efficient relative to its fully-paired counterpart and show that, as $\rho\to0$, paired DKPS degrades sharply while \PKPS does not.
The same representations support both benchmark prediction tasks and \PKPS-based methods outperform the baselines on both. 

\paragraph{Practical deployment.}
Our results suggest a simple workflow for practical deployment: (i) maintain the
cached responses (and, optionally, per-response scores) produced when
evaluating a model; (ii) fit \PKPS on the cached responses; and (iii) predict with a simple
regression in the perspective space, optionally ensembled with the sample score.
As shared caches like the Every Eval Ever datastore grow, each new upload
improves the perspective space, and the cumulative savings grow with them.

\paragraph{Limitations and future work.}
(i) \textit{Stochastic scoring functions.}
Our analysis assumes the mapping from a response to its score is deterministic.
Many modern evaluation frameworks use LLM-as-judge scoring \citep{zheng2023judging},
where scores are stochastic and may depend on contextual factors beyond the
individual response.
If each score takes the form $s(f(q))+\epsilon$ with $\epsilon$ i.i.d., repeated
scoring recovers the current setting in the limit and the extension is conceptually
straightforward.
(ii) \textit{Beyond uniform observation.}
Assumption~\ref{ass:noise} models observed cells as uniform subsamples, and our
experiments induce sparsity and unpairedness by uniform masking.
Missingness in deployed evaluation data is not uniform: which models are evaluated
on which benchmarks correlates with developer, release date, and expected
performance.
Extending the analysis to modeled observation processes is a natural and important next step.
(iii) \textit{Unpaired monitoring in multi-agent systems.}
Paired DKPS has been used as a tool for tracking the perspectives of multi-agent
systems \citep{helm2024tracking}.
These approaches require asking each agent a collection of fixed queries each time
point and are prohibitively expensive at scale.
The \PKPS removes this requirement as agents can be monitored through the
heterogeneous interactions they naturally produce.


\subsection*{AI use statement}

In this work, we used generative AI tools for implementing methods and simple data processing.
We also used them in the creation of artifacts (e.g., using Matplotlib to create figures) and to draft / edit parts of the manuscript with heavy user guidance.
We have reviewed all AI-assisted work. 
For example, generated code was verified and tested for correctness by all authors.
Generated text was heavily edited.
We take responsibility for the final content of this work,
including text, claims or artifacts produced with the aid of generative AI.


\subsubsection*{Acknowledgments}
We would like to thank Heather G. Patsolic and Brandon Duderstadt for helpful discussions throughout the development of this manuscript.
We gratefully acknowledge funding from the Defense Advanced Research Projects Agency (DARPA) Artificial Intelligence Quantified (AIQ) award number HR00112520026.

This project was born out of discussions within the EvalEval Coalition
(\raisebox{-0.2ex}{\includegraphics[height=1em]{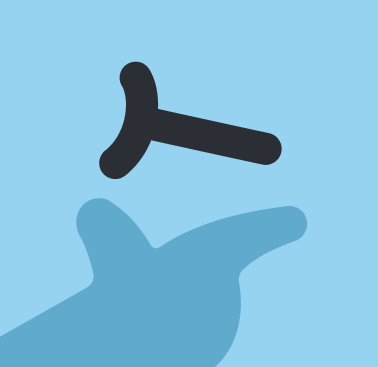}}\hspace{0.3em}\url{https://evalevalai.com/}),
and the authors are grateful to Brian Hu, Leshem Choshen, and Andreas Haupt for their early feedback.

\bibliographystyle{plainnat}
\bibliography{references}

\appendix
\section{Experimental details}
\label{app:details}

\paragraph{Suite.} $18$ tasks over four datasets from HELM-Lite \citep{helm}: MATH ($7$
subjects), WMT-14 ($5$ translation directions), MedQA, and LegalBench ($5$ subsets),
restricted to the $93$ models evaluated on all four. Table~\ref{tab:helmsuite} lists
every task with its query count; the loader subsamples each task to at most $120$
queries with a fixed seed, so datasets contribute comparably.

\begin{table}[h]
\centering
\caption{HELM suite composition: $18$ tasks over four datasets, with the number of unique
queries per task. Every task is scored for all $93$ shared models; the loader caps each
task at $120$ queries (seeded subsample).}
\label{tab:helmsuite}
\footnotesize
\begin{tabular}{llr @{\hspace{2em}} llr}
\toprule
Dataset & Task & Queries & Dataset & Task & Queries \\
\midrule
MATH & algebra & 135 & WMT-14 & cs--en & 873 \\
MATH & counting \& probability & 39 & WMT-14 & de--en & 776 \\
MATH & geometry & 38 & WMT-14 & fr--en & 754 \\
MATH & intermediate algebra & 52 & WMT-14 & hi--en & 387 \\
MATH & number theory & 30 & WMT-14 & ru--en & 613 \\
MATH & prealgebra & 86 & MedQA & med\_qa & 1000 \\
MATH & precalculus & 57 & LegalBench & abercrombie & 95 \\
LegalBench & corporate lobbying & 490 & LegalBench & citizenship questions & 1000 \\
LegalBench & function of decision section & 367 & LegalBench & proa & 95 \\
\bottomrule
\end{tabular}
\end{table}

\paragraph{Model coverage (HELM).} The $93$ shared models span $20$ developers:
{\footnotesize\ttfamily\raggedright 01-ai\_yi-34b, 01-ai\_yi-6b, 01-ai\_yi-large-preview,
AlephAlpha\_luminous-base, AlephAlpha\_luminous-extended, AlephAlpha\_luminous-supreme,
ai21\_j2-grande, ai21\_j2-jumbo, ai21\_jamba-1.5-large, ai21\_jamba-1.5-mini,
ai21\_jamba-instruct, allenai\_olmo-7b, amazon\_nova-lite-v1, amazon\_nova-micro-v1,
amazon\_nova-pro-v1, anthropic\_claude-2.0, anthropic\_claude-2.1,
anthropic\_claude-3-5-haiku-20241022, anthropic\_claude-3-5-sonnet-20240620,
anthropic\_claude-3-5-sonnet-20241022, anthropic\_claude-3-haiku-20240307,
anthropic\_claude-3-opus-20240229, anthropic\_claude-3-sonnet-20240229,
anthropic\_claude-instant-1.2, anthropic\_claude-instant-v1, anthropic\_claude-v1.3,
cohere\_command, cohere\_command-light, cohere\_command-r, cohere\_command-r-plus,
databricks\_dbrx-instruct, deepseek-ai\_deepseek-llm-67b-chat, deepseek-ai\_deepseek-v3,
google\_gemini-1.0-pro-001, google\_gemini-1.0-pro-002, google\_gemini-1.5-flash-001,
google\_gemini-1.5-flash-002, google\_gemini-1.5-pro-001, google\_gemini-1.5-pro-002,
google\_gemini-1.5-pro-preview-0409, google\_gemini-2.0-flash-exp, google\_gemma-2-27b-it,
google\_gemma-2-9b-it, google\_gemma-7b, google\_text-bison@001, google\_text-unicorn@001,
meta\_llama-2-13b, meta\_llama-2-70b, meta\_llama-2-7b, meta\_llama-3-70b, meta\_llama-3-8b,
meta\_llama-3.1-405b-instruct-turbo, meta\_llama-3.1-70b-instruct-turbo,
meta\_llama-3.1-8b-instruct-turbo, meta\_llama-3.2-11b-vision-instruct-turbo,
meta\_llama-3.2-90b-vision-instruct-turbo, meta\_llama-3.3-70b-instruct-turbo,
meta\_llama-65b, microsoft\_phi-2, microsoft\_phi-3-medium-4k-instruct,
microsoft\_phi-3-small-8k-instruct, mistralai\_mistral-7b-instruct-v0.3,
mistralai\_mistral-7b-v0.1, mistralai\_mistral-large-2407, mistralai\_mistral-medium-2312,
mistralai\_mixtral-8x22b, mistralai\_mixtral-8x7b-32kseqlen,
mistralai\_open-mistral-nemo-2407, nvidia\_nemotron-4-340b-instruct,
openai\_gpt-3.5-turbo-0613, openai\_gpt-4-0613, openai\_gpt-4-1106-preview,
openai\_gpt-4-turbo-2024-04-09, openai\_gpt-4o-2024-05-13, openai\_gpt-4o-2024-08-06,
openai\_gpt-4o-mini-2024-07-18, openai\_text-davinci-002, openai\_text-davinci-003,
qwen\_qwen1.5-110b-chat, qwen\_qwen1.5-14b, qwen\_qwen1.5-32b, qwen\_qwen1.5-72b,
qwen\_qwen1.5-7b, qwen\_qwen2-72b-instruct, qwen\_qwen2.5-72b-instruct-turbo,
qwen\_qwen2.5-7b-instruct-turbo, snowflake\_snowflake-arctic-instruct, tiiuae\_falcon-40b,
tiiuae\_falcon-7b, upstage\_solar-pro-241126, writer\_palmyra-x-004, writer\_palmyra-x-v2,
writer\_palmyra-x-v3.\par}

\paragraph{Representations.} All queries and responses are embedded with
\texttt{gemini-embedding-001} ($3072$-d). MATH and WMT-14 responses are free text and
are embedded directly. MedQA and LegalBench responses are single-token answers
(multiple-choice letters and class indices); each is standardized to the text it
denotes before embedding---the option text the letter selects, recovered from the
MedQA source data and verified against model behavior on $999$ of $1000$ instances,
and the class name the index encodes. Queries share one PCA space (capped at $64$
components) and responses another, both fit on observed rows only; the response
space's dimension is set to the total that separate per-dataset reductions would
retain (second Zhu--Ghodsi elbow \citepapp{zhu2006automatic} per dataset, at most $48$ components each; $92$ on
this suite), and response vectors are unit-normalized. Nothing separates the datasets
by construction: the query kernel, whose bandwidth is set from within-dataset
distances, determines how much weight cross-dataset comparisons receive.

\paragraph{Kernels.} The query kernel is an RBF whose bandwidth is selected per run by
leave-one-model-out $K$-NN ($K{=}8$) on the observed scores, over the grid
$\sigma\in\{0.03,0.1,0.3,1,3\}\times\mathrm{median}$ pairwise query distance plus the
near-delta (DKPS) limit; Appendix~\ref{app:sens} shows the cross-validated bandwidth
tracks the best fixed choice. The response kernel is linear on real data and RBF on
synthetic. Each (model, query) pair uses one sampled response ($r{=}1$).

\paragraph{Model embedding.} Models are embedded by classical MDS
(Eq.~\eqref{eq:mds}) with $d{=}8$ throughout, following \citet{dkps_qeff} and fixed a
priori ($d$ = latent dimension on synthetic); Appendix~\ref{app:sens} probes the
choice of $d$.

\paragraph{Regressor.} All embedding methods use a distance-weighted $K$-NN ($K{=}5$)
regression on the model embedding, fit \emph{leave-one-family-out}: all models from the target's developer are
excluded from its training set, so predictions never lean on near-duplicate siblings
\citep{dkps_qeff}. Held-out cells are predicted directly; observed cells leave their own
target out. For query-efficiency the regression targets are the reference models'
full-task scores. For completion the targets are residuals after a logit transform,
per-task standardization to unit variance, and a two-way (model $+$ task) additive bias;
this pipeline is identical to the LRMC baseline's, so the two methods differ
only in how the residual is modeled ($K$-NN on the perspective versus a low-rank
factorization) and performance differences are attributable to the representation.
Matching the per-task standardization is important: without it, the model offset is
dominated by the tasks with the largest logit variance.

\paragraph{Baselines.} Low-rank matrix completion (LRMC) follows BenchPress \citep{benchpress}: logit
space, per-task standardization, two-way bias, rank-$2$ ALS with ridge $\lambda{=}0.1$,
averaged over $5$ random initializations; on observed cells it is cross-fit so a cell's
prediction never sees its own sample. IRT is a 1PL (Rasch) model \citepapp{rasch1960} on binary tasks: item
difficulties by maximum likelihood (L-BFGS, \citealpapp{liu1989limited}) on the largest fully-observed
(model $\times$ item) block, per-model ability by bounded scalar MLE, clipped to
$[-4,4]$. The sample baseline is the cell's own $m$-query mean.

\paragraph{Ensemble.} On observed cells, the prediction is $\alpha\,\hat
y_{\mathrm{sample}}+(1-\alpha)\,\hat y_{\PKPS}$ with a scalar $\alpha$ per held-out
family, selected on a $21$-point grid by leave-one-family-out validation against the
other families' cached full scores (the same information the regressor trains on); on
missing cells, \PKPS and LRMC are blended with a weight fit by held-out cross-validation
against observed scores. No quantity ever uses the scores of the family being predicted.

\paragraph{Sweeps and reporting.}\begin{sloppypar} All real-data sweeps use $16$ seeds; curves and table
entries report the per-seed mean over tasks, then the mean ($\pm1$ SEM) over seeds. Query
efficiency: budgets $m\in\{1,2,4,8,16,32\}$ at full cohort and coverage; cohort
$n\in\{10,20,40,60,93\}$ (HELM) or $\{10,20,40,45\}$ (EEE) and task coverage
$p_{\mathrm{task}}\in\{0.2,0.35,0.5,0.7,0.9\}$, both at the few-query point $m{=}2$.
Completion: task coverage as above and query depth
$p_{\mathrm{query}}\in\{0.1,0.25,0.5,0.75,1.0\}$ with cohort lines $n\in\{10,\text{full}\}$;
task count $T\in\{4,8,12,18\}$ (HELM) or $\{4,8,12,16\}$ (EEE); the cohort sweep holds
$p_{\mathrm{task}}{=}p_{\mathrm{query}}{=}0.5$.\end{sloppypar}

\paragraph{Statistical tests.} Paired differences between methods are assessed with
two-sided Wilcoxon signed-rank tests \citepapp{wilcoxon1945individual} on per-task error differences: for each task we
average the per-cell absolute-error difference over models and seeds, and test the
per-task means against zero ($18$ tasks on HELM, $16$ on EEE).
The query-efficiency tests compare the ensemble to the sample score at $m{=}1$ (full
cohort and coverage); the completion tests compare the ensemble to LRMC at the
central operating point ($p_{\mathrm{task}}{=}p_{\mathrm{query}}{=}0.5$) at each
cohort size.
The per-cell improvement fractions quoted in Section~\ref{sec:results} are computed
from the same per-cell records, pooled over the conditional-figure sweeps.

\paragraph{Synthetic study.} $50$ seeds per configuration; latent dimension $5$, observed
embedding dimension $20$, response noise $0.5$; both kernels RBF.
Query-efficiency panels (Figure~\ref{fig:synthetic}a--e): every observed cell carries
$m$ queries and a sample score with measurement noise $4/\sqrt{m}$; the true
score is predicted by a distance-weighted average of the $K{=}8$ nearest models'
sample scores, restricted to cells they observe.
Completion panel (f): score noise $0.1$, held-out cells predicted by $K$-NN regression
with $K{=}5$.
Unswept levers default to $n{=}60$, $T{=}12$, $m{=}4$, $\rho{=}0$,
$p_{\mathrm{task}}{=}1$; panel-specific settings are given in each figure title.

\paragraph{Compute.} All experiments run on a single multi-core CPU node; text embeddings
are precomputed once through the embedding API and cached. Embedding the EEE suite
($30{,}304$ pooled responses plus $7{,}665$ queries, ${\sim}11$M tokens) took ${\sim}13$
minutes and under \$$2$ of API credit; each $16$-seed sweep completes in minutes (EEE) to
hours (HELM) on the same node.

\section{The Every Eval Ever suite}
\label{app:eee}

\begin{figure}[t]
\centering
\includegraphics[width=\textwidth]{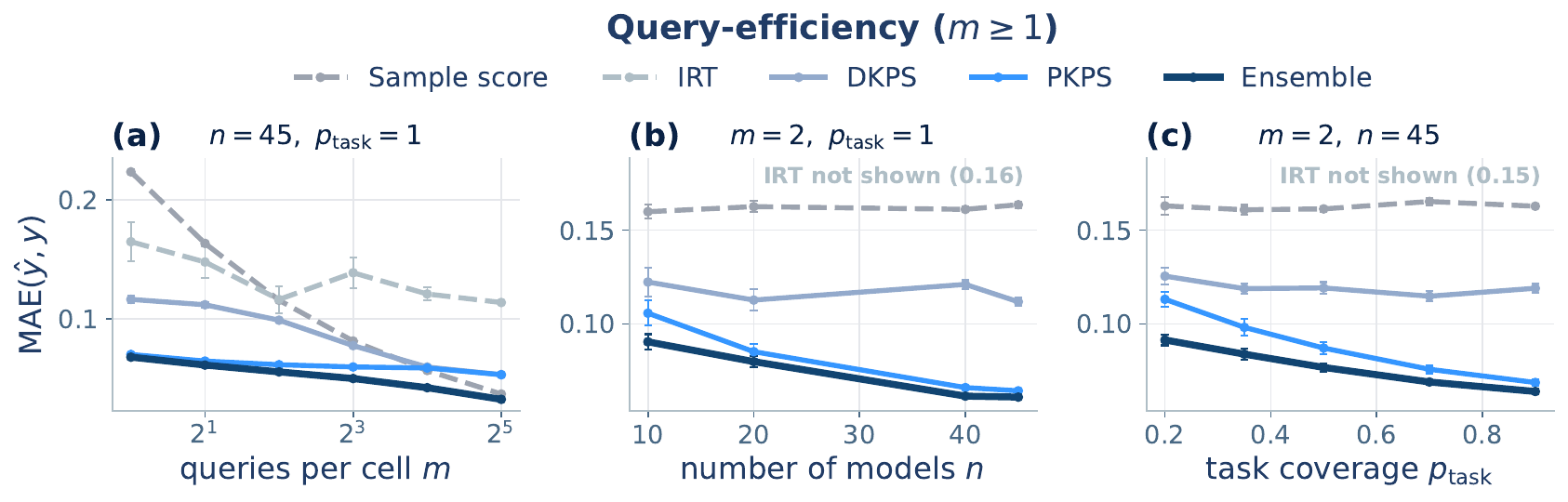}
\caption{Query-efficient evaluation on the Every Eval Ever suite ($45$ models, $16$
tasks; MAE against the true full-evaluation score, $\pm1$ SEM over $16$ seeds). The protocol
matches Figure~\ref{fig:qe}. The Ensemble is the most accurate at every budget and
\PKPS at every budget up to $m{=}8$; DKPS trails because the query pools are
unpaired, and IRT is poorly identified.}
\label{fig:eeeqe}
\end{figure}

\begin{figure}[t]
\centering
\includegraphics[width=\textwidth]{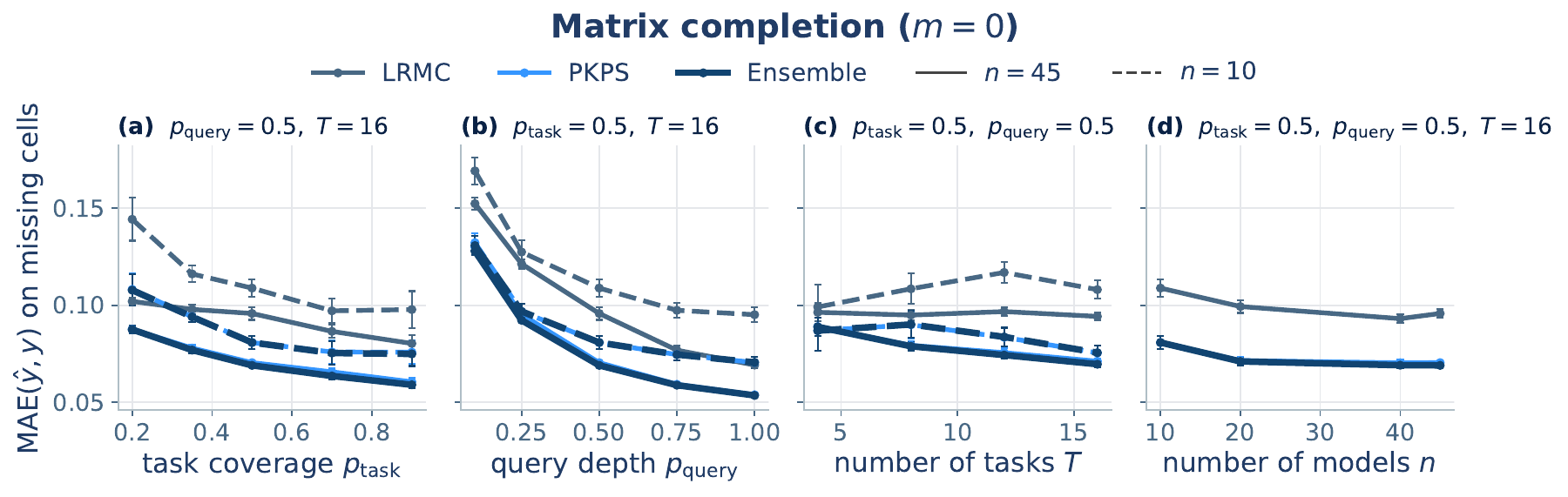}
\caption{Score completion on the Every Eval Ever suite (MAE on missing cells, $\pm1$
SEM; solid $n{=}45$, dashed $n{=}10$). The protocol matches
Figure~\ref{fig:completion}. Unlike on the HELM suite, \PKPS and the Ensemble
outperform LRMC at every operating point.}
\label{fig:eeecomp}
\end{figure}

The second suite is built from the Every Eval Ever datastore \citep{eee2026}, an
MIT-licensed, community-maintained collection of item-level evaluation results in a
unified format (inputs, raw model outputs, and per-item scores). The full store
covers $36$ benchmarks and $165$ models with only $15\%$ of (model, benchmark)
pairs evaluated (median $5$ benchmarks per model), and $59\%$ of model pairs share
no benchmark. Within a benchmark, $26$ of the $33$ with at least two models are
essentially fully paired ($23$ with identical item sets across every model pair),
while the seven arena-style benchmarks (six Fibble arenas and Wordle arena) are
natively unpaired: mean per-pair overlap $0.27$--$0.66$, median zero on six of the
seven, and per-model coverage of the item union $0.15$--$0.33$---in interactive
evaluation, the queries a model sees depend on its matchups. We take the five benchmarks with the
widest item-level model coverage---MATH-MC, GSM-MC, GPQA-Diamond, JudgeBench, and
RewardBench~2---restricted to the $45$ models present on all five, for which every
model answered every item; sub-tasks (MATH-MC
levels $1$--$5$; four JudgeBench and five RewardBench~2 categories) give $16$ tasks
(Table~\ref{tab:eeesuite}). Every domain is free text (including the
judging benchmarks' rationales), so all responses and queries are embedded with
\texttt{gemini-embedding-001} directly, and the response space is built exactly as in
Appendix~\ref{app:details}: one shared PCA over all five benchmarks (dimension $83$,
matched to the per-benchmark total), unit-normalized, with the query kernel gating
cross-benchmark comparisons. Each cell's observable pool is $32$ queries drawn with a
model-dependent seed, so two models share only incidental queries (median
$\rho_{ii'}^{(t)}=0.09$, mean $0.18$, largest on the smallest tasks); the
query-efficiency budget $m$ and the completion depth $p_{\mathrm{query}}$ subsample this
pool, and the prediction target is always the cell's full-depth mean over all items, which
is never embedded.
All scores are binary, so IRT applies to every task (on the HELM suite it excludes WMT).
Figures~\ref{fig:eeeqe} and \ref{fig:eeecomp} repeat the lever sweeps of
Figures~\ref{fig:qe} and \ref{fig:completion} on this suite; Table~\ref{tab:suite}
summarizes.

\begin{table}[h]
\centering
\caption{EEE suite composition: $16$ tasks over five benchmarks, with the number of items
per (model, task) cell. Every task is fully answered by all $45$ shared models; each
cell's observable pool is $32$ items, sampled per model.}
\label{tab:eeesuite}
\footnotesize
\begin{tabular}{llr @{\hspace{2em}} llr}
\toprule
Benchmark & Task & Items & Benchmark & Task & Items \\
\midrule
MATH-MC & level 1 & 430 & JudgeBench & coding & 42 \\
MATH-MC & level 2 & 882 & JudgeBench & knowledge & 154 \\
MATH-MC & level 3 & 1115 & JudgeBench & math & 56 \\
MATH-MC & level 4 & 1199 & JudgeBench & reasoning & 98 \\
MATH-MC & level 5 & 1288 & RewardBench~2 & factuality & 475 \\
GSM-MC & gsm\_mc & 1319 & RewardBench~2 & focus & 495 \\
GPQA-Diamond & gpqa\_diamond & 198 & RewardBench~2 & math & 183 \\
RewardBench~2 & precise if & 160 & RewardBench~2 & safety & 450 \\
\bottomrule
\end{tabular}
\end{table}

\paragraph{Model coverage (EEE).} The $45$ models shared by all five benchmarks span $13$
developers:
{\footnotesize\ttfamily\raggedright cohere/c4ai-command-a-03-2025, cohere/c4ai-command-r-08-2024,
cohere/c4ai-command-r-plus-08-2024, cohere/c4ai-command-r7b-12-2024,
cohere/command-a-reasoning-08-2025, deepseek/deepseek-r1-0528,
deepseek/deepseek-v3-1-terminus, deepseek/deepseek-v3-2, deepseek/deepseek-v3-2-speciale,
deepseek/deepseek-v4-flash-fp8, deepseek/deepseek-v4-pro, google/gemini-3-1-pro-preview,
google/gemma-2-27b-it, google/gemma-2-9b-it, google/gemma-3-12b-it, google/gemma-3-27b-it,
google/gemma-4-e2b-it, google/gemma-4-e4b-it, llm360/k2-v2-instruct,
meta/llama-4-maverick-17b-128e-instruct-fp8, meta/meta-llama-3-1-8b-instruct,
minimax/minimax-m2-5, mistral/mistral-small-4-119b-2603, moonshot/kimi-k2-5,
moonshot/kimi-k2-6, moonshot/kimi-k2-thinking, openai/gpt-5-5, openai/gpt-5-mini,
openai/gpt-5-nano, openai/gpt-oss-120b, openai/gpt-oss-20b,
qwen/qwen3-30b-a3b-thinking-2507, qwen/qwen3-5-122b-a10b, qwen/qwen3-5-2b,
qwen/qwen3-5-397b-a17b, qwen/qwen3-6-35b-a3b, qwen/qwen3-6-plus,
qwen/qwen3-next-80b-a3b-thinking, stepfun/step-3-5-flash, xiaomi/mimo-v2-flash,
zai-org/glm-4-6-fp8, zai-org/glm-4-7-flash, zai-org/glm-4-7-fp8, zai-org/glm-5-1-fp8,
zai-org/glm-5-fp8.\par}

\section{Sensitivity analyses}
\label{app:sens}

\begin{figure}[t]
\centering
\includegraphics[width=\textwidth]{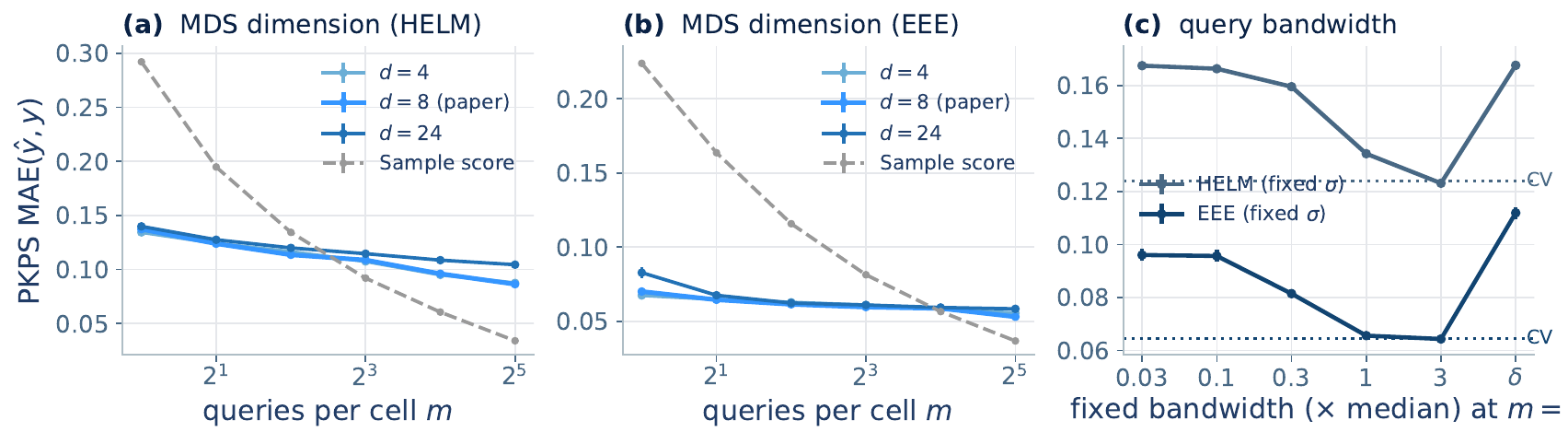}
\caption{Sensitivity analyses on the query-efficiency experiment ($\pm1$ SEM, $16$
seeds). \textbf{(a, b)} \PKPS MAE against budget at MDS dimension $d\in\{4,8,24\}$
(the paper's choice $d{=}8$ drawn heavier; sample dashed); $d{=}4$ and $d{=}8$ are
nearly indistinguishable on both suites, while $d{=}24$ trails at the extremes.
\textbf{(c)} MAE at $m{=}2$ with the query bandwidth fixed at each grid
multiplier ($\delta$ $=$ exact matching); dotted lines mark the cross-validated
selection, which matches the best fixed choice on both suites.}
\label{fig:sens}
\end{figure}

Figure~\ref{fig:sens} probes the two representation hyperparameters on the
query-efficiency experiment ($16$ seeds throughout). 

\textbf{MDS dimension (a, b).}
Repeating the budget sweep at $d\in\{4,8,24\}$, the $d{=}4$ and $d{=}8$ curves are
within $0.003$ of each other at every budget on both suites (at the smallest budgets,
smaller $d$ is very slightly better---fewer coordinates to estimate from one noisy
query per cell), while $d{=}24$ costs up to $0.013$ at the smallest EEE budgets and up
to $0.018$ at the largest HELM budgets; the paper's choice ($d{=}8$, following
\citealp{dkps_qeff}) is never the worst point at any budget. 

\textbf{Query bandwidth (c).} Fixing the
bandwidth at each grid multiplier instead of cross-validating traces the expected curve:
near-delta multipliers and the exact-matching limit ($k_Q(u,u')=\mathbf{1}\{u=u'\}$,
abbreviated $\delta$) are substantially worse
(unpaired pools leave them little to compare), broad multipliers are best, and the cross-validated choice matches the best fixed multiplier on both suites
($0.124$ vs.\ $0.123$ on HELM, $0.065$ vs.\ $0.064$ on EEE)---the one hyperparameter
that matters is the one the method selects automatically. Sensitivity to the number of
reference models is part of the main experiments (the cohort sweeps of
Figures~\ref{fig:qe}b and \ref{fig:completion}d).

\textbf{Pipeline ablations.} Repeating the budget sweep with one choice changed at a
time: the cross-validated ensemble weight is best or tied at every budget on both
suites (the inverse-variance alternative loses $0.007$ on HELM and $0.012$ on EEE
at $m{=}1$); replacing the $K$-NN regressor with least squares helps slightly on
HELM ($0.132$ vs.\ $0.137$ at $m{=}1$) but hurts on EEE ($0.061$ vs.\ $0.053$ at
$m{=}32$); an RBF response kernel changes \PKPS by at most $0.002$; and enforcing
exactly disjoint EEE query pools (true $\rho=0$) changes \PKPS by at most $0.006$
at $m\le8$.

\section{Proofs of Theorems~\ref{prop:conc} and \ref{thm:qeff}}
\label{app:proof}

\paragraph{Formal definitions and assumptions.} We state the formal statements referenced in Section~\ref{sec:method}.
\begin{definition}[Population \PKPS]
\label{def:pop}
Let each model $i$'s (task, query, response) triples be drawn i.i.d.\ from a design
distribution $P_i$: each triple independently draws a task, then a query conditioned on
the task, then the model's response. Write $z=(u,x_i(u))$ for an embedded
query--response pair (response noise included). The \emph{population affinity} is
\[
A^*_{ii'}\;=\;\frac{\mathbb{E}\,[k_Q(u,u')\,k_R(x_i(u),x_{i'}(u'))]}{\mathbb{E}\,[k_Q(u,u')]},
\qquad z\sim P_i,\; z'\sim P_{i'} \text{ independent},
\]
the population squared dissimilarities are $D^{*2}(i,i')=A^*_{ii}+A^*_{i'i'}-2A^*_{ii'}$, and
the \emph{population perspectives} $\psi_1,\dots,\psi_n\in\mathbb{R}^d$ (the rows of
$\Psi^*\in\mathbb{R}^{n\times d}$) are the rank-$d$ spectral embedding of the
double-centered matrix $B^*=-\tfrac12 J D^{*2} J$, where $J=I-\tfrac1n\mathbf{1}\mathbf{1}^\top$:
$\Psi^*=[v^*_1,\dots,v^*_d]\operatorname{diag}(\lambda_1(B^*)^{1/2},\dots,\lambda_d(B^*)^{1/2})$
for the eigenpairs of the $d$ largest eigenvalues of $B^*$. $B^*$ need not be positive
semidefinite.
\end{definition}
\begin{assumption}[Observation model]\label{ass:noise}
Each scoring function is the mean of bounded per-query scores, and each observed cell
$Q_{i}^{(t)}$ is a uniform subsample of $Q^{(t)}$ at the budget $m$.
\end{assumption}
\begin{assumption}[Boundedness]\label{ass:bound}
Kernels and embedded responses are bounded.
\end{assumption}
\begin{assumption}[Population overlap]\label{ass:overlap}
The \emph{cross-model query-kernel mass} is bounded below:
$\kappa=\inf_{i,i'}\mathbb{E}[k_Q(u,u')]>0$.
\end{assumption}
\begin{assumption}[Spectral identifiability]\label{ass:spec}
The $d$-th largest eigenvalue of $B^*$ is positive and separated from the next:
$\lambda_d(B^*)>0$ and $\Delta=\lambda_d(B^*)-\lambda_{d+1}(B^*)>0$.
\end{assumption}
\begin{assumption}[Pairing deficiency]\label{ass:gap}
The cross-model mass of the exact-matching kernel vanishes:
$\kappa_{0}=\sup_{i\neq i'}\mathbb{E}[k_{0}(u,u')]=0$ for independent queries
$u\sim P_{i}$, $u'\sim P_{i'}$, where $k_{0}(u,u')=\mathbf{1}\{u=u'\}$;
equivalently, no two query designs share an atom.
\end{assumption}
\begin{assumption}[Independent caches]\label{ass:iid}
The models and their query designs $(f_{i},P_{i})$, $i=0,\dots,n$, are i.i.d., and
given them the caches are drawn independently across models.
\end{assumption}
Under Assumption~\ref{ass:iid}, each model together with its cache,
$Z_{i}=(f_{i},C_{i})$, is an i.i.d.\ draw from a law $\nu_{m}$ that depends on the
budget $m$.
\begin{definition}[Fixed-budget representation]\label{def:info}
For $\sigma>0$ let $h_{\sigma}(z,z')=A(z,z)+A(z',z')-2A(z,z')$, where $A(z,z')$ is
Eq.~\eqref{eq:A} evaluated on the caches of $z$ and $z'$, and let
\[
b_{\sigma}(z,z')=-\tfrac12\big(h_{\sigma}(z,z')-\bar h_{\sigma}(z)-\bar h_{\sigma}(z')+\bar{\bar h}_{\sigma}\big),
\quad
\bar h_{\sigma}(z)=\!\int\! h_{\sigma}(z,w)\,d\nu_{m}(w),
\quad
\bar{\bar h}_{\sigma}=\!\int\!\bar h_{\sigma}\,d\nu_{m}.
\]
The integral operator $(T_{\sigma}g)(z)=\int b_{\sigma}(z,w)g(w)\,d\nu_{m}(w)$ on
$L^{2}(\nu_{m})$ is self-adjoint and compact. Let $\theta_{1}\ge\dots\ge\theta_{d}$ be
its $d$ largest eigenvalues with orthonormal eigenfunctions $e_{1},\dots,e_{d}$, and
$\theta_{d+1}$ the supremum of the rest of its spectrum. The \emph{fixed-budget
representation} is
$\phi_{m,\sigma}(z)=\big(\theta_{k}^{-1/2}\int b_{\sigma}(z,w)e_{k}(w)\,d\nu_{m}(w)\big)_{k=1}^{d}$,
the continuous analogue of Eq.~\eqref{eq:mds} \citepapp{trosset2024continuous}, and
$I_{m}(\sigma)=\mathrm{Var}\big(\mathbb{E}[y_{0}\mid\phi_{m,\sigma}(Z_{0})]\big)$.
\end{definition}
\begin{assumption}[Fixed-budget spectral identifiability]\label{ass:specm}
For $T_{\sigma^{\star}}$, $\theta_{d}>0$ and $\theta_{d}>\theta_{d+1}$.
\end{assumption}

\paragraph{Population quantities.} Throughout, the quantities of
Definition~\ref{def:pop} are used as stated there. By Assumption~\ref{ass:bound} the
kernels are bounded, $k_Q\in[0,1]$ after normalization (e.g., RBF) and $|k_R|\le B_R$,
and by Assumption~\ref{ass:spec} the rank-$d$ spectral embedding exists with
$\lambda_d(B^*)>0$.

\begin{lemma}[Affinity concentration]
\label{lem:conc}
Under Assumptions~\ref{ass:bound} and \ref{ass:overlap}, for models $i,i'$ with
$\min(|Q_i|,|Q_{i'}|)\ge m$ and any $\delta\in(0,1)$ we have, with probability at
least $1-\delta$,
\[
|A_{ii'}-A^*_{ii'}|\;\le\;\frac{2(1+B_R)}{\kappa^2}\,t_m(\delta),
\qquad t_m(\delta)=\bar B\sqrt{\frac{\log(4/\delta)}{2m}},\quad \bar B=\max(2B_R,1),
\]
provided $t_m(\delta)\le\kappa/2$.
\end{lemma}
\begin{proof}
\begin{sloppypar}
The numerator $N=\frac{1}{|Q_i||Q_{i'}|}\sum_{j,l}k_Q(u_j,u'_l)\,k_R(x_{ij},x_{i'l})$ and
denominator $Z=\frac{1}{|Q_i||Q_{i'}|}\sum_{j,l}k_Q(u_j,u'_l)$ of Eq.~\eqref{eq:A} are
two-sample $U$-statistics of order $(1,1)$ in the i.i.d.\ pairs $z_j\sim P_i$,
$z'_l\sim P_{i'}$, with kernels of range at most $2B_R$ and $1$ respectively (the
response draw is part of $z$, so no per-query response replication is needed). By
Hoeffding's inequality for two-sample $U$-statistics
\citepapp[Section~5]{hoeffding1963}, each deviates from its mean by more than its range times
$\sqrt{\log(4/\delta)/(2m)}$ with probability at most $\delta/2$; since $\bar B$ dominates
both ranges, a union bound gives $|N-N^*|\le t_m(\delta)$ and $|Z-Z^*|\le t_m(\delta)$
jointly with probability $\ge1-\delta$. On this event, since $Z^*\ge\kappa$ by Assumption~\ref{ass:overlap} and $t_m(\delta)\le\kappa/2$,
we have $Z\ge\kappa/2$, and
\[
|A_{ii'}-A^*_{ii'}|=\frac{|NZ^*-N^*Z|}{ZZ^*}
\;\le\;\frac{|N-N^*|}{Z}+\frac{|N^*|\,|Z-Z^*|}{ZZ^*}
\;\le\;\frac{2\,t_m(\delta)}{\kappa}+\frac{2B_R\,t_m(\delta)}{\kappa^2}
\;\le\;\frac{2(1+B_R)}{\kappa^2}\,t_m(\delta),
\]
using $|N^*|\le B_R$ and $\kappa\le1$.
\end{sloppypar}
\end{proof}

\begin{lemma}[Perspective perturbation]
\label{lem:pert}
Let $\varepsilon_\infty=\max_{i,i'}|A_{ii'}-A^*_{ii'}|$ and
$\beta=\|B^*\|_{\mathrm{op}}/\Delta$. Under Assumption~\ref{ass:spec} there is an
orthogonal $W\in\mathbb{R}^{d\times d}$ such that
\[
\max_i\;\|\hat\psi_i-W\psi_i\|\;\le\;
\frac{53\,(1+\beta)\,n\,\varepsilon_\infty}{\sqrt{\lambda_d(B^*)}}\,.
\]
\end{lemma}
\begin{proof}
Entrywise, $|D^2-D^{*2}|\le4\varepsilon_\infty$, and double centering inflates an
entrywise bound by at most a factor of $4$ (each entry of $JMJ$ is a signed combination
of an entry, a row mean, a column mean, and the grand mean of $M$), so $E=B-B^*$
satisfies $\|E\|_{\mathrm{op}}\le\|E\|_F\le8\,n\,\varepsilon_\infty$. Let $\hat P$ and
$P^*$ be the orthogonal projections onto the eigenvectors of the $d$ largest
eigenvalues of $B$ and $B^*$. Then $\Psi^*\Psi^{*\top}=P^*B^*P^*$, and
$\hat\Psi\hat\Psi^\top$ differs from $\hat PB\hat P$ only in that negative eigenvalues
among the top $d$ are set to zero. Each such eigenvalue satisfies
$|\lambda_k(B)|\le|\lambda_k(B)-\lambda_k(B^*)|$ because $\lambda_k(B^*)\ge\lambda_d(B^*)>0$,
so by the Hoffman--Wielandt inequality \citepapp{hoffman1953variation}
$\|\hat\Psi\hat\Psi^\top-\hat PB\hat P\|_F\le\|E\|_F$. Writing
\[
\hat\Psi\hat\Psi^\top-\Psi^*\Psi^{*\top}
=\big(\hat\Psi\hat\Psi^\top-\hat PB\hat P\big)+\hat PE\hat P
+(\hat P-P^*)B^*\hat P+P^*B^*(\hat P-P^*),
\]
and using the Davis--Kahan bound
$\|\hat P-P^*\|_F\le2\sqrt2\,\|E\|_F/\Delta$ \citep[Theorem~2]{yu2015useful}, we obtain
\[
\|\hat\Psi\hat\Psi^\top-\Psi^*\Psi^{*\top}\|_F
\;\le\;\Big(2+\frac{4\sqrt2\,\|B^*\|_{\mathrm{op}}}{\Delta}\Big)\|E\|_F
\;\le\;6\,(1+\beta)\,\|E\|_F .
\]
Since $\sigma_d(\Psi^*)^2=\lambda_d(B^*)$, the Procrustes alignment lemma of
\citetapp[Lemma~5.4]{tu2016procrustes} gives
\[
\min_{W\in O(d)}\;\|\hat\Psi-\Psi^*W\|_F^2
\;\le\;\frac{\|\hat\Psi\hat\Psi^\top-\Psi^*\Psi^{*\top}\|_F^2}{2(\sqrt2-1)\,\lambda_d(B^*)}\,.
\]
Combining, $\max_i\|\hat\psi_i-W\psi_i\|\le\|\hat\Psi-\Psi^*W\|_F\le
\frac{6(1+\beta)}{\sqrt{2(\sqrt2-1)}}\cdot\frac{8n\varepsilon_\infty}{\sqrt{\lambda_d(B^*)}}
\le\frac{53(1+\beta)\,n\varepsilon_\infty}{\sqrt{\lambda_d(B^*)}}$.
\end{proof}

\begin{proof}[Proof of Theorem~\ref{prop:conc}]
Apply Lemma~\ref{lem:conc} at
confidence $\delta/n^2$ to each of the at most $n^2$ model pairs; a union bound gives,
with probability at least $1-\delta$,
\[
\varepsilon_\infty\;=\;\max_{i,i'}|A_{ii'}-A^*_{ii'}|
\;\le\;\frac{2(1+B_R)\bar B}{\kappa^{2}}\sqrt{\frac{\log(4n^{2}/\delta)}{2m}}.
\]
Plugging this into Lemma~\ref{lem:pert} yields
$\max_i\|\hat\psi_i-W\psi_i\|\le
C\kappa^{-2}\lambda_d^{-1/2}(1+\beta)\,n\sqrt{\log(4n^{2}/\delta)/2m}$ with
$C=106(1+B_R)\bar B$, which depends only on the kernel bounds. Lemma~\ref{lem:conc}
requires $t_m(\delta/n^2)\le\kappa/2$; in the complementary case the claimed bound
already exceeds the trivial one---every perspective satisfies
$\|\psi_i\|^2\le\lambda_1(B)\le 8B_Rn$ and likewise for $\hat\psi_i$, so
$\max_i\|\hat\psi_i-W\psi_i\|\le6\sqrt{B_Rn}$, while the right-hand side is at least
$C\kappa^{-1}(1+\beta)\sqrt{n}/(2\bar B\sqrt{8B_R})\ge6\sqrt{B_Rn}$ once $C$ also dominates
$96B_R\bar B$---so the theorem holds for every $m$ after enlarging $C$ by a constant
factor.
\end{proof}

The remainder of this appendix proves Theorem~\ref{thm:qeff}. The first claim is a
cross-validation transfer argument with a guarded selection rule. The second claim
combines a lower bound on the MSE of every paired estimator under
Assumption~\ref{ass:gap} with the large-$n$ limit of the representation at the fixed
budget $m$ (Lemmas~\ref{lem:nys} and~\ref{lem:info}).

\begin{proof}[Proof of Theorem~\ref{thm:qeff}(i)]
Split the reference models into an embedding set $\mathcal{L}$, a regression set $\mathcal{R}$, and a
validation set $\mathcal{V}$, with $|\mathcal{L}|=\ell$ and $|\mathcal{V}|=n_v$. For each $\sigma$ in the finite
grid $\Sigma\ni0$, compute the classical MDS embedding (Eq.~\eqref{eq:mds}) of $\mathcal{L}$,
with top eigenpairs $(\hat\lambda_k,\hat v_k)_{k\le d}$ of $B_{\mathcal{L}}$, and embed any model
$z$ by the Nystr\"om out-of-sample extension \citepapp{bengio2004out} (see also \citealp{trosset2008out})
\begin{equation}
\hat\phi_{\mathcal{L}}(z)=\Big(\hat\lambda_k^{-1/2}\sum_{j\in\mathcal{L}}\hat v_k(j)\,\hat b_{\mathcal{L}}(z,j)\Big)_{k=1}^{d},
\qquad
\hat b_{\mathcal{L}}(z,j)=-\tfrac12\Big(D^2(z,j)-\bar D^2_{\mathcal{L}}(z)-\bar D^2_{\mathcal{L}}(j)+\bar{\bar D}{}^2_{\mathcal{L}}\Big),
\label{eq:nys}
\end{equation}
where $\bar D^2_{\mathcal{L}}(z)=\ell^{-1}\sum_{j\in\mathcal{L}}D^2(z,j)$,
$\bar{\bar D}{}^2_{\mathcal{L}}=\ell^{-1}\sum_{j\in\mathcal{L}}\bar D^2_{\mathcal{L}}(j)$, and
$\hat\lambda_k^{-1/2}$ is read as $0$ when $\hat\lambda_k\le0$. For $z\in\mathcal{L}$,
$\hat b_{\mathcal{L}}(z,\cdot)$ is the corresponding row of $B_{\mathcal{L}}$, so Eq.~\eqref{eq:nys}
reproduces the classical MDS coordinates. Let $\hat y_{\sigma}$ be $K$-nearest-neighbor
regression on $\mathcal{R}$'s pairs $(\hat\phi_{\mathcal{L}}(f_i),y_i)$, with validation loss
$\hat L(\sigma)=\tfrac1{n_v}\sum_{i\in\mathcal{V}}(\hat y_{\sigma}(f_i)-y_i)^2$.
Let $\tau=\sqrt{\log(2|\Sigma|/\delta')/2n_v}$ be the validation confidence
radius. The guarded selection is
\[
\hat\sigma=\begin{cases}
\arg\min_{\sigma\in\Sigma}\hat L(\sigma) & \text{if }\min_{\sigma}\hat L(\sigma)\le\hat L(0)-2\tau,\\[2pt]
0 & \text{otherwise.}
\end{cases}
\]
Conditional on $\mathcal{L}$ and $\mathcal{R}$, the validation losses are i.i.d.\ across the draw of
validation models and bounded in $[0,1]$, so by Hoeffding's inequality \citepapp{hoeffding1963} and a
union bound over $\Sigma$, with probability at least $1-\delta'$,
$\sup_{\sigma\in\Sigma}|\hat L(\sigma)-L(\sigma)|\le\tau$, where $L(\sigma)$ is
the MSE on a fresh target conditional on $\mathcal{L}$ and $\mathcal{R}$. On this event, if
$\hat\sigma\ne0$ then
\[
L(\hat\sigma)\;\le\;\hat L(\hat\sigma)+\tau\;\le\;\hat L(0)-2\tau+\tau\;\le\;L(0),
\]
and if $\hat\sigma=0$ the two estimators coincide; either way
$\mathrm{MSE}(\hat h_{n}^{(K)})\le\mathrm{MSE}(h_{n,0}^{(K)})$, since the
$\sigma=0$ member coincides with paired DKPS on the shared queries with the same
regressor (Section~\ref{sec:qeff}). This holds at every $(m,\rho)$.
\end{proof}

\begin{lemma}[Fixed-budget limit of the representation]
\label{lem:nys}
Fix $m$ and $\sigma=\sigma^{\star}$, and suppose Assumptions~\ref{ass:bound},
\ref{ass:iid}, and~\ref{ass:specm} hold. As $\ell\to\infty$ there are orthogonal
$W_{\mathcal{L}}\in\mathbb{R}^{d\times d}$, depending on $\mathcal{L}$, such that
$\|\hat\phi_{\mathcal{L}}(Z)-W_{\mathcal{L}}\phi_{m,\sigma}(Z)\|\to0$ in probability for $Z\sim\nu_m$
independent of $\mathcal{L}$.
\end{lemma}
\begin{proof}
Write $h=h_\sigma$, $b=b_\sigma$, $\phi=\phi_{m,\sigma}$, and let $Z_j$, $j\in\mathcal{L}$, be
the embedding models with their caches, i.i.d.\ with law $\nu_m$ by
Assumption~\ref{ass:iid}. By Assumption~\ref{ass:bound}, $|A|\le B_R$, so
$|h|\le4B_R$ and $|b|\le8B_R$. We use the versions
$e_k(z)=\theta_k^{-1}\int b(z,w)e_k(w)\,d\nu_m(w)$, which satisfy
$|e_k|\le8B_R/\theta_k$ and $\phi_k=\theta_k^{1/2}e_k$.

\emph{Centering.} By Hoeffding's inequality and a union bound,
$\max_{j\in\mathcal{L}}|\bar D^2_{\mathcal{L}}(j)-\bar h(Z_j)|$, $|\bar{\bar D}{}^2_{\mathcal{L}}-\bar{\bar h}|$, and,
for any fixed $z$, $|\bar D^2_{\mathcal{L}}(z)-\bar h(z)|$ are $O_P(\sqrt{\log\ell/\ell})$. Hence
$\max_{j\in\mathcal{L}}|\hat b_{\mathcal{L}}(z,j)-b(z,Z_j)|\to0$ in probability for every fixed $z$, and
$B_{\mathcal{L}}$ differs from $[b(Z_i,Z_j)]_{i,j\in\mathcal{L}}$ by $o_P(\ell)$ in operator norm.

\emph{Spectrum and eigenvectors.} Let $E_{\mathcal{L}}\in\mathbb{R}^{\ell\times d}$ have rows
$(e_1(Z_j),\dots,e_d(Z_j))$ and $\hat U=[\hat v_1,\dots,\hat v_d]$. By the
random-matrix approximation of integral operators
\citepapp{koltchinskii1998asymptotics,koltchinskii2000random}, together with Weyl's
inequality \citepapp{weyl1912asymptotische} and the Davis--Kahan theorem \citepapp{davis1970rotation} for the $o_P(\ell)$ centering perturbation,
$\hat\lambda_k/\ell\to\theta_k$ for $k\le d$, and, since
$\theta_1,\dots,\theta_d$ are isolated from the rest of the spectrum
(Assumption~\ref{ass:specm}), the spectral projections converge:
$\|\hat U\hat U^\top-\ell^{-1}E_{\mathcal{L}}E_{\mathcal{L}}^\top\|_F\to0$ in probability. With
$\ell^{-1}E_{\mathcal{L}}^\top E_{\mathcal{L}}\to I_d$, this gives orthogonal $W_{\mathcal{L}}$ with
$\ell^{-1/2}\|\ell^{1/2}\hat U-E_{\mathcal{L}}W_{\mathcal{L}}^\top\|_F\to0$. Applying the same argument to
each distinct value among $\theta_1,\dots,\theta_d$ lets $W_{\mathcal{L}}$ be block diagonal on
the corresponding eigenspaces, so it commutes with
$\Theta=\mathrm{diag}(\theta_1,\dots,\theta_d)$.

\emph{Fixed $z$.} Write
$\hat\phi_{\mathcal{L}}(z)=(\hat\Lambda/\ell)^{-1/2}\,\ell^{-1/2}\hat U^\top\hat\beta(z)$ with
$\hat\Lambda=\mathrm{diag}(\hat\lambda_1,\dots,\hat\lambda_d)$ and
$\hat\beta(z)_j=\hat b_{\mathcal{L}}(z,j)$, and let $\beta(z)_j=b(z,Z_j)$ and
$e=(e_1,\dots,e_d)$. Then
\begin{align*}
\ell^{-1/2}\hat U^\top\hat\beta(z)-W_{\mathcal{L}}\!\int\! b(z,w)e(w)\,d\nu_m(w)&
=\\ & \tfrac1\ell\big(\ell^{1/2}\hat U-E_{\mathcal{L}}W_{\mathcal{L}}^\top\big)^{\!\top}\hat\beta(z)
\\ &+W_{\mathcal{L}}\tfrac1\ell E_{\mathcal{L}}^\top\big(\hat\beta(z)-\beta(z)\big)
\\ &+W_{\mathcal{L}}\Big(\tfrac1\ell E_{\mathcal{L}}^\top\beta(z)-\!\int\! b(z,w)e(w)\,d\nu_m(w)\Big).
\end{align*}
By Cauchy--Schwarz the first term is at most
$\|\hat\beta(z)\|_\infty\,\ell^{-1/2}\|\ell^{1/2}\hat U-E_{\mathcal{L}}W_{\mathcal{L}}^\top\|_F\to0$; the
second is at most $\sqrt d\max_k\|e_k\|_\infty\max_j|\hat\beta(z)_j-\beta(z)_j|\to0$
by the centering step; the third is an average of $\ell$ i.i.d.\ bounded vectors
minus its mean and vanishes by the law of large numbers. Since
$(\hat\Lambda/\ell)^{-1/2}\to\Theta^{-1/2}$ and $W_{\mathcal{L}}$ commutes with $\Theta$,
$\hat\phi_{\mathcal{L}}(z)-W_{\mathcal{L}}\phi(z)\to0$ in probability for every fixed $z$. For
$Z\sim\nu_m$ independent of $\mathcal{L}$, dominated convergence gives
$\Pr(\|\hat\phi_{\mathcal{L}}(Z)-W_{\mathcal{L}}\phi(Z)\|>\epsilon)\to0$ for every $\epsilon>0$.
\end{proof}

\begin{lemma}[The representation retains its information]
\label{lem:info}
Under the conditions of Lemma~\ref{lem:nys}, let
$g_{\mathcal{L}}(x)=\mathbb{E}[y_0\mid\hat\phi_{\mathcal{L}}(Z_0)=x,\,\mathcal{L}]$ and
$\hat I_{\mathcal{L}}=\mathrm{Var}\big(g_{\mathcal{L}}(\hat\phi_{\mathcal{L}}(Z_0))\mid \mathcal{L}\big)$. Then for every
$\epsilon>0$, $\Pr\big(\hat I_{\mathcal{L}}\ge I_m(\sigma)-\epsilon\big)\to1$ as $\ell\to\infty$.
\end{lemma}
\begin{proof}
Write $X=\phi_{m,\sigma}(Z_0)$, $\hat X=\hat\phi_{\mathcal{L}}(Z_0)$,
$g(x)=\mathbb{E}[y_0\mid X=x]$, and $v=\mathrm{Var}(y_0)$; scores lie in $[0,1]$, and
$(Z_0,y_0)$ is independent of $\mathcal{L}$. For any bounded measurable $q$,
$\mathbb{E}[(y_0-q(\hat X))^2\mid \mathcal{L}]\ge\mathbb{E}[(y_0-g_{\mathcal{L}}(\hat X))^2\mid \mathcal{L}]=v-\hat I_{\mathcal{L}}$.
Fix $\epsilon'>0$ and choose a bounded continuous $q_0$ with
$\mathbb{E}[(g(X)-q_0(X))^2]\le\epsilon'^2$, which is possible because bounded
continuous functions are dense in $L^2$ of the law of $X$, and set
$q(x)=q_0(W_{\mathcal{L}}^\top x)$. By the triangle inequality in $L^2$ conditional on $\mathcal{L}$,
\[
\big(v-\hat I_{\mathcal{L}}\big)^{1/2}\;\le\;\|y_0-g(X)\|+\|g(X)-q_0(X)\|+r_{\mathcal{L}}
\;\le\;\big(v-I_m(\sigma)\big)^{1/2}+\epsilon'+r_{\mathcal{L}},
\]
where $r_{\mathcal{L}}^2=\mathbb{E}[(q_0(X)-q_0(W_{\mathcal{L}}^\top\hat X))^2\mid \mathcal{L}]$ and we used
$\|y_0-g(X)\|^2=v-I_m(\sigma)$. Since $W_{\mathcal{L}}^\top\hat X-X\to0$ in probability
(Lemma~\ref{lem:nys}) and $q_0$ is bounded and continuous,
$q_0(W_{\mathcal{L}}^\top\hat X)-q_0(X)\to0$ in probability, so $\mathbb{E}[r_{\mathcal{L}}^2]\to0$ by
bounded convergence and $r_{\mathcal{L}}\to0$ in probability. Squaring and letting
$\epsilon'\to0$ gives the claim.
\end{proof}

\paragraph{Contrast with paired DKPS.} Under the limit kernel $k_{0}$
($k_{0}(u,u')>0$ iff $u=u'$), only identical query pairs enter the double sum, so the
$U$-statistic runs over the $\rho m$ shared queries
and the rate in Lemma~\ref{lem:conc} becomes $O_P\big((\rho m)^{-1/2}\big)$: DKPS's
guarantee requires $\rho m\to\infty$ and vacates under Assumption~\ref{ass:gap},
where the cross-model mass $\kappa_{0}$ is zero and its distance is undefined
(Section~\ref{sec:method}).

\begin{proof}[Proof of Theorem~\ref{thm:qeff}(ii)]
\emph{Floor.} Under Assumption~\ref{ass:gap}, $\Pr(q_j=q'_l)=0$ for every pair of
independently drawn queries, so almost surely every identity-matched sum is
empty: any measurable completion of a $\sigma=0$ estimator predicts as a function
of the reference data alone, independent of the target. Hence, conditional on the
reference data,
$\mathrm{MSE}(h_{n,0}^{(K)})=v+(\mathbb{E}y_{0}-\hat y)^2\ge v$
(for every neighbor count, every split, and every completion).

\emph{Achievability.} Fix $0<c<I_m(\sigma^\star)$, let
$\epsilon=(I_m(\sigma^\star)-c)/4$, and let $\ell$, $|\mathcal{R}|$, and $n_v$ grow
proportionally with $n$. Conditional on $\mathcal{L}$, the regression pairs
$(\hat\phi_{\mathcal{L}}(Z_i),y_i)$, $i\in\mathcal{R}$, and the target's $(\hat X,y_0)$ with
$\hat X=\hat\phi_{\mathcal{L}}(Z_0)$ are i.i.d.\ (Assumption~\ref{ass:iid}), so for
$K_n\to\infty$ with $K_n/n\to0$, $K_n$-nearest-neighbor regression with ties broken
at random is $L^2$-consistent for $g_{\mathcal{L}}$ \citepapp{stone1977consistent}. The cross term
vanishes because $\mathbb{E}[y_0-g_{\mathcal{L}}(\hat X)\mid\hat X,\mathcal{L}]=0$, so
\[
L(\sigma^\star)=\mathbb{E}\big[(y_0-g_{\mathcal{L}}(\hat X))^2\mid \mathcal{L}\big]
+\mathbb{E}\big[(\hat y_{\sigma^\star}-g_{\mathcal{L}}(\hat X))^2\mid \mathcal{L},\mathcal{R}\big]
=v-\hat I_{\mathcal{L}}+o_P(1),
\]
and Lemma~\ref{lem:info} gives $L(\sigma^\star)\le v-I_m(\sigma^\star)+2\epsilon$
with probability tending to one. Choose $n_v$ large enough that
$4\tau\le I_m(\sigma^\star)-2\epsilon$ and $\tau\le\epsilon/2$. On this event and
the deviation event of part (i), the guard triggers, since
$\hat L(\sigma^\star)\le L(\sigma^\star)+\tau\le v-I_m(\sigma^\star)+2\epsilon+\tau
\le v-3\tau\le\hat L(0)-2\tau$, using $\hat L(0)\ge L(0)-\tau\ge v-\tau$. Then
\[
L(\hat\sigma)\le\hat L(\hat\sigma)+\tau\le\hat L(\sigma^\star)+\tau
\le L(\sigma^\star)+2\tau\le v-I_m(\sigma^\star)+3\epsilon<v-c
\le\mathrm{MSE}(h_{n,0}^{(K_n)})-c .
\]
The margin $c>0$ supplies the strict clause of Definition~\ref{def:qeff} at the
fixed budget $m$.
\end{proof}

\bibliographystyleapp{plainnat}
\bibliographyapp{references}

\end{document}